\documentclass{article}
\usepackage{multirow}
\usepackage{PRIMEarxiv}
\usepackage{array}
\usepackage[utf8]{inputenc} 
\usepackage[T1]{fontenc}    
\usepackage{hyperref}       
\usepackage{url}            
\usepackage{booktabs}       
\usepackage{amsfonts}       
\usepackage{nicefrac}       
\usepackage{microtype}      
\usepackage{lipsum}
\usepackage{fancyhdr}       
\usepackage{graphicx}       
\graphicspath{{media/}}     
\usepackage{amsmath}
\usepackage{algorithm}
\usepackage{algpseudocode}
\usepackage{array}
\title{Disrupting Model Merging: A Parameter-Level Defense Without Sacrificing Accuracy}

\author{
 Wei Junhao \thanks{Equal contribution.} \\
 RIKEN AIP ; Institute of Science Tokyo \\
 China\\
 \texttt{wei.j.ac@m.titech.ac.jp}\\
 \And
  Yu Zhe \footnotemark[1]\\
  RIKEN AIP \\
  China \\
  \texttt{zhe.yu@riken.jp} \\
  \And
  Jun Sakuma \\
  RIKEN AIP; Institute of Science Tokyo\\
  Japan\\
  \texttt{sakuma@c.titech.ac.jp} \\}

\date{}
    
\begin{document}

\bibliographystyle{plain}
\maketitle
\begin{abstract}
Model merging is a technique that combines multiple finetuned models into a single model without additional training, allowing a free-rider to cheaply inherit specialized capabilities. This study investigates methodologies to suppress unwanted model merging by free-riders. Existing methods such as model watermarking or fingerprinting can only detect merging in hindsight. In contrast, we propose a first proactive defense against model merging. Specifically, our defense method modifies the model parameters so that the model is disrupted if the model is merged with any other model, while its functionality is kept unchanged if not merged with others. Our approach consists of two modules, rearranging MLP parameters and scaling attention heads, which push the model out of the shared basin in parameter space, causing the merging performance with other models to degrade significantly. We conduct extensive experiments on image classification, image generation, and text classification to demonstrate that our defense severely disrupts merging while retaining the functionality of the post-protect model. Moreover, we analyze potential adaptive attacks and further propose a dropout-based pruning to improve our proposal's robustness. 
\end{abstract}

\section{Introduction}
\label{sec:intro}

The pretrain–finetune paradigm trains models on large datasets before finetuning on smaller, task-specific ones, enabling efficient adaptation to specialized tasks. For instance, CNNs like ResNet \cite{resnet2016he} pretrained on ImageNet \cite{imagenet2009deng} extract rich visual features for downstream tasks. Platforms like Hugging Face have further popularized this approach, making it fundamental to language \cite{brown2020languagemodelsfewshotlearners, touvron2023llamaopenefficientfoundation}, vision \cite{dosovitskiy2021an}, and multimodal learning \cite{radford2021learningtransferablevisualmodels, rombach2021highresolution}.

The success of this paradigm has led to model merging \cite{Survery_ModelMerging_2024}, a technique that integrates multiple fine-tuned models from the same pretrained source without additional training, often by linearly combining parameters. Specifically, consider two models: a digit classification model and an object classification model. The merged model is capable of handling both digit and object classification tasks. A basic model merging technique is parameter averaging, where the parameters of two fine-tuned models are combined using weight averaging \cite{utans1996weight}. More advanced methods, such as Task Arithmetic \cite{ilharco2023editing} and TIES-Merging \cite{yadav2023tiesmerging}, further refine the averaging approach. With nearly 30,000 merged models available on Hugging Face, model merging has become a practical and efficient approach for customizing models across diverse applications.

\begin{figure*}
    \centering    \includegraphics[width=\linewidth]{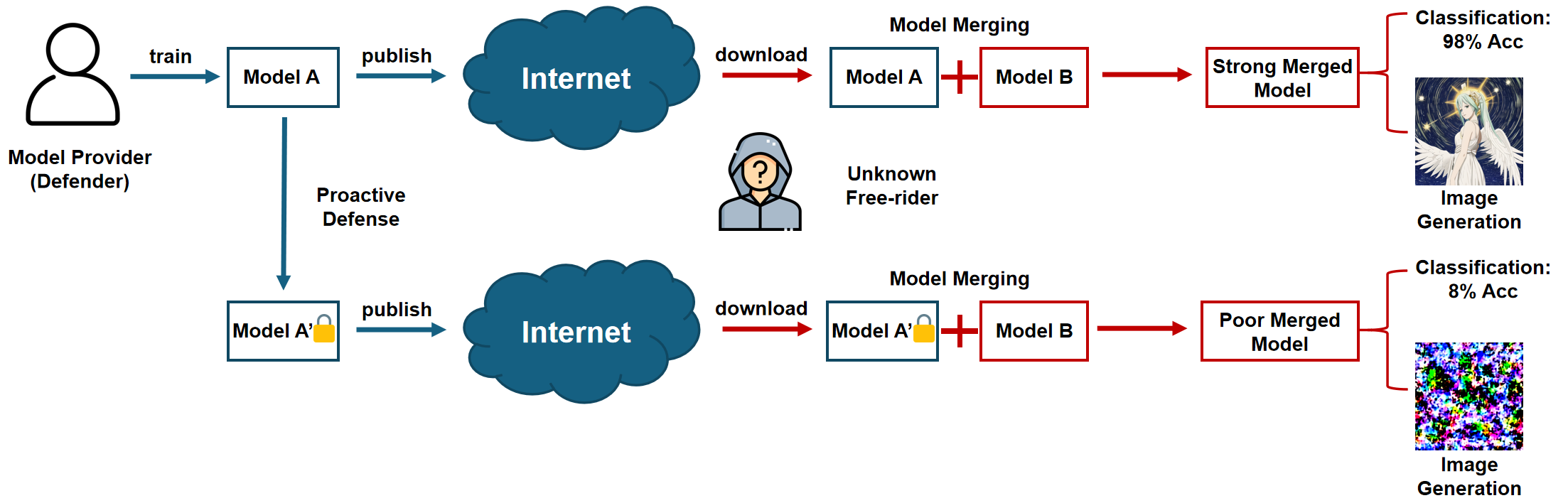}
    \caption{Illustration of proactive defense against model merging}
    \label{fig: story}
\end{figure*}

Although model merging is convenient and low-cost, it introduces potential risks that have received relatively little attention. \cite{Cong2024haveyoumergemymodel, yamabe2024mergeprintrobustfingerprintingmerging} argue that a model’s capabilities constitute valuable intellectual property (IP), and unwanted merging could lead to IP infringement. For example, suppose a company open-source finetuned models to showcase their technological advancements under certain license terms. Then, a free-rider could merge such models into their own finetuned versions at minimal cost by ignoring the license terms, which threatens IP rights. This infringement occurs because the merged model retains the specialized performance from the original finetuned models without requiring the same level of computational investment or expertise (Figure \ref{fig: story} top). Unlike traditional software, where direct code reuse is easily traceable, model merging blends parameter spaces in a way that makes it difficult to attribute contributions to specific sources. As a result, a merged model can inherit proprietary capabilities while obfuscating its origins, effectively allowing free-riders to bypass licensing restrictions and benefit from intellectual efforts they did not contribute to. This not only diminishes the commercial value of the original model but also raises concerns about fair attribution and ethical usage, as free-riders can publicly deploy the merged model under the guise of independent innovation. Moreover, since merging does not inherently degrade performance and can even enhance it in some cases, it incentivizes free-riders to exploit this loophole, further exacerbating the risk of IP infringement.

To address this, \cite{Cong2024haveyoumergemymodel, yamabe2024mergeprintrobustfingerprintingmerging} have proposed model watermarking or fingerprinting techniques that enable post-merging verification, allowing model owners to detect whether their models have been used without permission. Yet, these methods are inherently passive and do not prevent merging in the first place. This gap presents a critical security challenge, and there is an urgent need for proactive defenses that make model merging ineffective on protected models, preventing unwanted usage from the outset.

In this study, to mitigate the IP infringement risk posed by model merging, we ask a fundamental research question: \textbf{How can we proactively protect a finetuned model from unwanted merging while preserving its original capabilities?} An illustration of proactive defense is shown in Figure \ref{fig: story} bottom. We consider that a proactive defense should meet two core requirements. First, before merging, the protected model must retain its original performance since releasing a degraded model would defeat the purpose of showcasing its strengths. Second, after merging, the model’s performance should degrade significantly, thereby discouraging free-riders from exploiting it. 

To fulfill these pre-merging and post-merging requirements, we propose a proactive defense methods, termed \emph{\textbf{PaRaMS}}, short for \textbf{P}arameter re\textbf{a}rrangement \& \textbf{Ra}ndom \textbf{M}ulti-head \textbf{S}caling, applied to two fundamental modules common across modern machine learning models: the Multi-Layer Perceptron (MLP) layers and attention modules of Transformer-based architectures. PaRaMS perturbs model parameters without altering the model’s functionality, thereby preserving its original performance before merging. At the same time, these parameter modifications shift the protected model away from other finetuned models that share the same pretrained checkpoint, causing merged models to suffer poor performance. Thus, PaRaMS meets both requirements of retaining pre-merging capabilities and undermining post-merging utility. Moreover, since most model merging methods operate on MLP \cite{utans1996weight}, Attention \cite{sundar2024multimodal}, or both \cite{ilharco2023editing,yadav2023tiesmerging}, our approach is generalizable to most merging techniques.

We summarize our main contributions as follows:

\begin{enumerate}
\item We are the first to propose a proactive solution against the free-rider threat in model merging, addressing the security gap where a free-rider can cheaply inherit a defender’s finetuned capabilities.
    
\item Our method preserves the pre-merging model’s functionality while significantly decreasing the performance of post-merging models.
    
\item Through experiments on multiple tasks and datasets, we show that our approach retains the model’s performance yet severely undermines merging performance, and we provide in-depth analysis to explain why it succeeds in disrupting model merging.

\item Considering the possibility of a free-rider attempting to bypass our defense, we comprehensively evaluated adaptive strategies to circumvent our method. In response to these adaptive approaches, we also proposed corresponding solutions, ensuring that the free rider either fails to bypass our method or incurs an excessively high cost to do so.
\end{enumerate}

\section{Preliminaries}
\textbf{Model Merging.} 
Model merging techniques allow the merged model to inherit each model’s task-specific performance without requiring additional training, relying instead on parameter-level operations. Formally, let $\theta$ be model parameters, with $\theta_{\mathrm{pre}}$ denoting pretrained parameters and $\theta_i$ finetuned parameters on dataset $\mathcal{D}_i$. For $n$ task-specific models $\{\theta_1, \dots, \theta_n\}$ derived from the same pretrained checkpoint $\theta_{\mathrm{pre}}$, model merging can be written as
\[
\theta_{\text{merge}} \;=\; \mathcal{M}\bigl(\theta_{\mathrm{pre}}, \theta_1, \dots, \theta_n\bigr),
\]
where $\mathcal{M}$ is a merging algorithm, often based on linear interpolation of parameters, as introduced later. Let $\mathcal{T}_i$ denote the data distribution (or test set) for the $i$-th task, and $\mathrm{Perf}: \theta \times \mathcal{T} \to \text{R}$ be a performance evaluation score of $\theta$ on samples from $\mathcal{T}_i$ (e.g. for classification task, this refers to accuracy). Under this notation, the goal of model merging is to find a merged model to retain each task’s performance:
\[
\mathrm{Perf}\bigl(\theta_{\mathrm{merge}}, \mathcal{T}_i\bigr)
\;\approx\;
\mathrm{Perf}\bigl(\theta_i, \mathcal{T}_i\bigr)
\quad
\forall\,i \in \{1,\dots,n\}.
\]
In this paper, we focus on model merging methods that rely on parameter-level operations, which cover most mainstream approaches.

\textbf{Task Arithmetic.}
Task Arithmetic (TA) \cite{ilharco2023editing} is a classic model merging method, and many approaches \cite{yadav2023tiesmerging,ortiz-jimenez2023task} are improvements on this method. Therefore, we will first introduce this method. In TA, they define a task vector as an element-wise difference: $\tau_i = \theta_i-\theta_{\mathrm{pre}}$. It then merges these task vectors by a linear combination:
$\theta_{\mathrm{merged}}=\theta_{\mathrm{pre}}+\lambda\Sigma_{i=1}^{n}\tau_i$, where $\lambda$ is the scaling coefficient. By directly adding task vectors of different tasks to the pretrained parameters, a multi-task model can be constructed. This multi-task model is known to be able to achieve a similar performance \cite{ortiz-jimenez2023task, zhou2024emergencecrosstasklinearitypretrainingfinetuning, Survery_ModelMerging_2024} of each individual model on corresponding task.

\textbf{Other Merging Techniques.}
Several subsequent works have focused on improving TA, similarly rely on additive parameter operations and require a same pretrained model. However, they further refine the merging coefficients (e.g., AdaMerging \cite{AdaMerging_ICLR_2024}) or adjust parameter subspaces (e.g., TIES-Merging \cite{yadav2023tiesmerging}, DARE \cite{yu2024language}). For more details, please see the appendix.

\section{Problem Formulation}
\subsection{Threat Model} 

\textbf{Attack Scenario.}
We consider a common workflow where a \textbf{defender} starts with a large pretrained model and then finetunes it on specialized data to optimize performance for a particular domain or task. Once finetuning is complete, the defender open-sources this finetuned model, often to demonstrate its capabilities to the broader community. Meanwhile, a \textbf{free-rider} obtain this open-sourced model with the intent of merging it with a model under their control. We suppose the free-rider's model is also finetuned from the same pretrained checkpoint, enabling straightforward parameter combination. This assumption is reasonable because most finetuning workflows in deep learning begin with widely used pretrained models such as CLIP, Vision Transformers (ViTs), Diffusion models or Llama, which serve as standard initialization points across different applications. By merging the two models, the free-rider aims to inherit the defender’s specialized performance with extremely low cost, and combine it with whatever capabilities already exist in their own model. To prevent such free-riders, the \textbf{defender} seeks to develop proactive defenses that maintain the model’s original performance while making its performance significantly degraded when merged with any other models.

\textbf{Free-rider's capability.}
We assume the free-rider has limited computing resources, making expensive training approaches such as knowledge distillation or full re-training impractical. Instead, the free-rider relies on model merging as a low-cost alternative to acquire the defender’s specialized capabilities. To enable effective merging, the free-rider possesses a finetuned model originating from the same pretrained checkpoint as the defender’s model and has white-box access to the defender’s open-source model. The success of the free-rider’s attack is measured by whether the merged model achieves a similar performance as the defender's model without sacrificing the ability to deal with the free rider’s task.

\textbf{Defender's capability.}
The defender has full control over their model’s parameters and can apply modifications to protect it. These modifications must preserve the model’s original performance while preventing effective merging. The defender does not have access to the free-rider’s model but assumes, in line with common model merging practices, that it originates from the same pretrained checkpoint; otherwise, merging would inherently fail. The defender's success is defined by ensuring that their model remains fully functional for legitimate users, while any merged model suffers significant performance degradation on the defender’s task.

\subsection{Problem Setup.}
Let $\theta_{\text{def}}$ and $\theta_{\text{fr}}$ be the defender's model and free-riders model derived from the same open-source pretrained model $\theta_{\text{pre}}$ finetuned using data $\mathcal{D}_{\text{def}}$ and $\mathcal{D}_{\text{fr}}$, respectively. The free-rider attempts to obtain $\theta_{\text{merge}} = \mathcal{M}(\theta_{\text{pre}}, \theta_{\text{def}}, \theta_{\text{fr}})$ which has comparable performance on both tasks, 
\begin{align*}
    \text{Perf}(\theta_{\text{merge}}, \mathcal{T}_{\text{def}}) &\approx \text{Perf}(\theta_{\text{def}}, \mathcal{T}_{\text{def}}), \\
    \text{Perf}(\theta_{\text{merge}},\mathcal{T}_{\text{fr}}) &\approx \text{Perf}(\theta_{\text{fr}}, \mathcal{T}_{\text{fr}}).
\end{align*}
But for the defender, being merged by unknown free-rider is not wanted thus they employ a proactive defense for its model $\eta: \theta \to \theta$ to modify parameters without changing model architecture and publish the modified model $\hat{\theta}_{\text{def}}=\eta(\theta_{\text{def}})$. $\hat{\theta}_{\text{def}}$ is required to keep functional equivalence, i.e.,
\begin{equation}
f(x, \theta_{\text{def}})= f(x, \hat{\theta_{\text{def}}}) \quad\text{for any}\,\, x, 
\label{eq:a}
\end{equation}
therefore, $\text{Perf}(\hat{\theta}_{\text{def}}, \mathcal{T})=\text{Perf}(\theta_{\text{def}}, \mathcal{T})$ holds for any dataset $\mathcal{T}$.  
The free-rider may use model merging to get a merged model $\hat{\theta}_{\text{merge}}=\mathcal{M}(\theta_{\text{pre}}, \hat{\theta}_{\text{def}}, \theta_{\text{fr}})$, and the defender wants that the performance of the merged model on $\mathcal{T}_{\text{def}}$ is significantly low:
\begin{equation}
   \text{Perf}(\hat{\theta}_{\text{merge}}, \mathcal{T}_{\text{def}})\ll\text{Perf}(\theta_{\text{merge}}, \mathcal{T}_{\text{def}}).
    \label{eq:b}
\end{equation}
In summary, our problem is to design a proactive defense $\eta$ that satisfies Eq. \ref{eq:b} (performance degradation after model merge) under the condition Eq. \ref{eq:a} (functional equivalence before model merge).

\section{Method}
In this section, we introduce our defense, termed as \emph{\textbf{PaRaMS}}, that satisfies the conditions defined in the Eq. \ref{eq:a} and \ref{eq:b}. PaRaMS employs two distinct parameter transformations satisfying functional equivalence, tailored for two common modules, MLP and Attention modules. In the next subsection, we introduce the idea to disrupt the model with preserving functional invariance.

\begin{figure}
    \centering
    \includegraphics[width=0.7\linewidth]{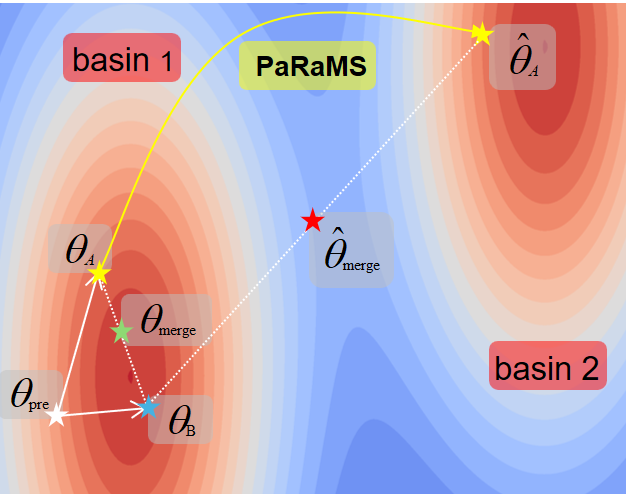}
    \caption{Loss landscape before and after applying our proposal}
    \label{fig: params}
\end{figure}
\subsection{Why Model Merging Works} \label{sec:insightonMM}
To disrupt model merging, we first ask why model merging works. In \cite{zhou2024emergencecrosstasklinearitypretrainingfinetuning}, they describe a basin in the loss landscape as a region where a model's parameters converge to a low-loss solution during optimization. They also found that, in the pretrain–finetune paradigm, finetuned models with the same pretrained model often end up in a shared or closely aligned basin (Figure \ref{fig: params}, basin 1). They then conclude this is the reason why model merging works. Considering this observation from the opposite perspective, it also suggests that if the two model parameters located in different distant basins, model merge would disrupt even if the models were derived from the same checkpoint (Figure \ref{fig: params}, basin 2). This can be experimentally confirmed in Section \ref{sec:exp} empirically. Our proactive defense against model merge is designed based on this observation.
More specifically, our idea is to use a model modification $\eta$ that satisfies the following two conditions: (1) The modified model $\eta(\theta)$ is functionally equivalent to the original model $\theta$, and (2) $\eta(\theta)$ are largely distant from $\theta$ in the model parameter space.

The first property guarantees that the performance of the model modified by our proactive defense is equivalent to that of the original model. In addition, even if model merging is performed, models modified by our proactive defense are expected to fail due to the second property. Figure \ref{fig: params} illustrates this idea.

\subsection{Our Proposal}\label{sec:method}
In this section, we propose two proactive defense methods that satisfy the two conditions introduced in the last subsection. The first is MLP parameter rearrangement, which is specifically designed for MLP architectures. The second is Random Multi-head Scaling, which is designed for multi-head attention blocks.

\textbf{MLP Parameter Rearrangement.}
\label{sec:mlp_permutation} In the following, we suppose the target model that consists of a 2-layer MLP for simplicity, but our method is applicable to any target model as long as it contains two or more consecutive MLP layers immediately. Concretely, consider a two-layer MLP of the following form:
\[
\mathrm{MLP}(X) \;=\; W_{2}\,\sigma\bigl(W_{1}\,X + b_{1}\bigr)\;+\;b_{2},
\]
where $\sigma$ is an element-wise nonlinear activation. We remark that element-wise nonlinear activations are commonly used for modern machine learning architectures (e.g., Sigmoid, ReLU \cite{2018arXiv180308375A}, tanh, etc.).

The idea is to randomly shuffle the weight matrices $W_1$ and $W_2$ of the MLP in an element-wise manner \cite{ainsworth2023git}. If the two shuffles are generated independently, this makes the behavior of the MLP completely different from the original model. In our construction, the shuffle provided for $W_2$ is set to restore the shuffle applied to $W_1$. To realize this, we introduce a permutation matrix $P$, which is a square binary matrix with exactly one “1” in each row and column and “0”s elsewhere. By multiplying this matrix with the MLP weight, we can reorder vector coordinates. Given such a matrix $P$, $P^{-1} = P^\mathsf{T}$ and element-wise non-linear function $\sigma$, we can find the following process does not change the MLP's output, as proved in \cite{ainsworth2023git}.
\[
W_{1}' = P\,W_{1},\quad b_{1}'=P\,b_1,\quad
W_{2}' = W_{2}\,P^\mathsf{T}.
\]
Here, $b_1$ refers to biases in the first layer of MLP, $W_1$ and $W_2$ are weights in the first/second layers of MLP.

The functional equivalence is satisfied for any random $P$, but considering that our goal is to make the model merge disrupt, the optimal permutation matrix is one that makes the distance between the model parameters as large as possible before and after the permutation. Formally, considering the realistic situation that a model has $n$ MLPs, we can apply the permutation $P_i$ and $P^\mathsf{T}_i$ to every $\text{MLP}_i$ ($i=1,...,n$). We then summarize all the operations of $P_i$ and $P^\mathsf{T}_i$  into a function $\eta_{\text{perm}}$, and our optimization objective is shown below:
\[
\arg\max_{\eta_{\text{perm}}} \bigl\|\theta^{\text{MLP}}_{\text{pre}} - \eta_{\text{perm}}(\theta^{\text{MLP}}_{\text{def}})\bigr\|^2.
\]
Here, $\theta^{\text{MLP}}_{\text{pre}}$ refers to all MLPs' parameters in the pretrained model. $\theta^{\text{MLP}}_{\text{def}}$ refers to all MLPs' parameters in the defender model. This optimization can be reformulated as a linear assignment problem, as proposed by \cite{ainsworth2023git} which attempts to optimize the permutation matrix to minimize two models' distance. We follow \cite{ainsworth2023git} to solve this optimization problem. More details about optimization are shown in the appendix.

\textbf{Random Multi-head Scaling.}\label{sec:attn_scaling} 
Next, we propose random multi-head scaling, which works as a proactive defense for attention block. An attention module with its corresponding output projection is defined as below:
\[
\mathrm{Attention} \;=\; \mathrm{softmax}\Bigl(\frac{Q\,K^\mathsf{T}}{\sqrt{d}}\Bigr)V W_{O},
\] where $Q, K, V$ is the query, key and value matrix, $W_O$ is the projection matrix. We leverage the fact that $QK^\mathsf{T} = QA(KA^{-1})^\mathsf{T}$ when A is real diagonal matrix. By introducing such random scaling with $A$, we can change the model parameter values while keeping functional equivalence. This can be applied to $ V W_{O}$ in the same manner. Similarly, for the value matrix \( V \) and its associated output projection matrix \( W_{O} \), we can apply the same transformation by multiplying \( V \) with a real diagnoal matrix $B$ and compensating by multiplying \( W_{O} \) with \(B^{-1}\), ensuring that the overall output of the attention module remains unchanged. Proof see the appendix.

Specifically, for each attention head $i$, we sample two diagonal matrices: $A_i, B_i$, whose diagonal entries are drawn from a uniform distribution $\mathcal{U}(s_{\min}, s_{\max})$ and update:
\begin{align*}
&Q_i \gets Q_i\,A_i,
\quad
K_i \gets K_i\,A_i^{-1} \\
V_i \gets &V_i \,B_i,
\quad
W_{O}[:, i] \gets B_i^{-1}\,W_{O}[:, i].
\end{align*}

We summarize all the scalings into a function $\eta_{\text{scaling}}$. Our method $\eta=\eta_{\text{perm}} \circ \eta_{\text{scaling}}$, which directly combines MLP parameter rearrangement and random multi-head scaling. The detailed algorithm is shown in the appendix. 

\section{Experiment}\label{sec:exp}

\subsection{Experiment Setup}

\textbf{Models.} In our experiments, we considered three tasks: image classification, image generation, and text classification. For image classification, we use three ViT \cite{dosovitskiy2021an} models: ViT-B-16, ViT-B-32, and ViT-L-14. These ViT models have been pre-trained within the CLIP \cite{radford2021learningtransferablevisualmodels} framework. For the image generation task, we use Stable Diffusion 1.5 (SD1.5) \cite{rombach2021highresolution}. For the text classification task, we use Llama2 \cite{touvron2023llama}. All models listed here contain MLPs and attention blocks.

\textbf{Datasets.} For image classification, we use seven datasets for the evaluations: Cars \cite{carsdataset}, RESISC45 \cite{resisc45}, SVHN \cite{SVHNdataset}, GTSRB \cite{gtsrbdataset}, MNIST \cite{LeCun2005TheMD}, EuroSAT \cite{helber2019eurosatnoveldatasetdeep} and DTD \cite{dtddataset}. Details about these datasets are shown in the appendix.

\textbf{Merging methods.} To demonstrate the generality of our method, we tested its effectiveness across various model merging approaches as Task arithemetic (TA) \cite{ilharco2023editing}, Weight Average (WA) \cite{pmlr-v162-wortsman22a},  TIES-Merging \cite{yadav2023tiesmerging} and AdaMerging (ADA) \cite{AdaMerging_ICLR_2024} along with a plug-in module DARE \cite{yu2024language}.  Details about these methods are shown in the appendix. 

\textbf{Evaluation strategy.}
In our experiments, we evaluate the performance of four distinct types of models based on two key aspects: merge situation and protection mechanisms. The first aspect defines whether a model is unmerged (specialized for a single task) or merged (a multi-task model capable of handling the free-rider model's task and the defender model's task). The second aspect defines whether a model is guarded with our protection or not. Specifically, we define the four models as follows: \textbf{Unmerged Model without Protection (UMP-)}; \textbf{Merged Model without Protection (MMP-)}; \textbf{Unmerged Model with Protection (UMP+)}; and \textbf{Merged Model with Protection (MMP+)}. For evaluation, unmerged models (UMP- and UMP+) are tested only on their designated tasks, while merged models (MMP- and MMP+) are assessed across all their supported tasks. Our preliminary goal is to ensure the performance of UMP+ are the same as UMP- and that the performance of MMP+ is much less than that of MMP-.

\subsection{Main Results}
\subsubsection{Results on Image classification}
\textbf{Comprehensive results on task arithmetic.}
In the first experiment, we trained both the defender’s and free-rider’s models on any two distinct datasets out of the seven, and measured the performance of protected and unprotected model before and after merge.
We denote the task corresponding to the defender model as $\mathcal{T}_{\text{def}}$ and the task corresponding to the free-rider model as $\mathcal{T}_{\text{fr}}$.  
Here, we employed ViT-B-32 for the models. Since ViT contains both MLP and multi-head attention, we applied MLP parameter rearrangement and random multi-head scaling for protection. 
We evaluated our proposal against TA because it serves as the foundation for many model merging methods. 

First, we measured the classification accuracy on $\mathcal{T}_{\text{def}}$ for four models: UMP-, UMP+, MMP-, MMP+. Moreover, for MMP- and MMP+, we additionally measured the classification accuracy on $\mathcal{T}_{\text{fr}}$. If PaRaMS works, performance of MMP+ on both tasks would be significantly degraded compared to MMP-. The results are summarized in Table~\ref{ta_vitb32_cpr}. We observe a huge gap between performance of MMP- and MMP+ on both $\mathcal{T}_{\text{def}}$ and $\mathcal{T}_{\text{fr}}$, achieving at least gap of 65.16\% ($67.82\% \to 2.66\%$ in GTSRB/DTD setting). Since our method does not alter the functionality of the defender model, UMP- and UMP+ have the same performance. 

\begin{figure}
    \centering 
    \includegraphics[width=1\linewidth]{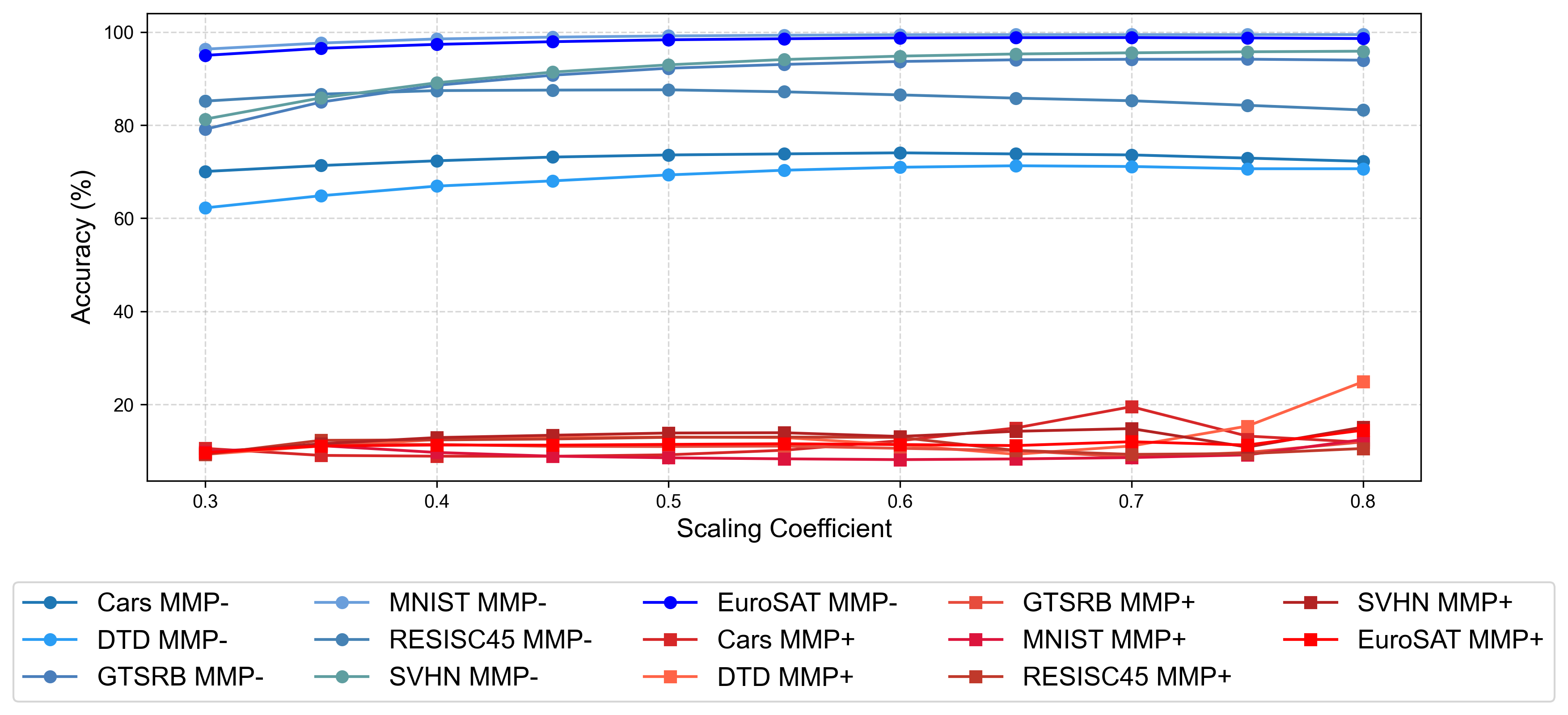}
    \caption{The classification accuracy of MMP-/MMP+ (merged by task arithmetic) with coefficient 0.3 to 0.8. Here, the defenser's model is trained on EuroSAT, and we have six different free-rider tasks. Here, model is ViT-B-32.}
    \label{fig: scaling coef}
\end{figure}

\begin{table*}[t]
\centering
\caption{The classification accuracy of MMP-/MMP+ (merged by TA) on $\mathcal{T}_{\text{def}}$ and $\mathcal{T}_{\text{fr}}$ on ViT-B-32 ($\lambda=0.8$).}
\label{ta_vitb32_cpr}
\resizebox{0.9\textwidth}{!}{%
\begin{tabular}{cc|ccccccc}
\hline
\multicolumn{2}{l|}{\multirow{2}{*}{\begin{tabular}[c]{@{}l@{}}MMP- Accuracy (\%) on $\mathcal{T}_{\text{def}}$/$\mathcal{T}_{\text{fr}}$\\ MMP+ Accuracy (\%) on $\mathcal{T}_{\text{def}}$/$\mathcal{T}_{\text{fr}}$\end{tabular}}} &
  \multicolumn{7}{c}{$\mathcal{T}_{\text{fr}}$} \\ \cline{3-9} 
\multicolumn{2}{l|}{} &
  Cars &
  RESISC45 &
  EuroSAT &
  SVHN &
  GTSRB &
  MNIST &
  DTD \\ \hline
\multicolumn{1}{c|}{\multirow{7}{*}{$\mathcal{T}_{\text{def}}$}} &
  Cars &
  \textit{NA} &
  \begin{tabular}[c]{@{}c@{}}70.29/94.24\\ 0.51/2.13\end{tabular} &
  \begin{tabular}[c]{@{}c@{}}72.23/99.76\\ 0.50/12.78\end{tabular} &
  \begin{tabular}[c]{@{}c@{}}67.53/97.23\\ 0.47/12.62\end{tabular} &
  \begin{tabular}[c]{@{}c@{}}67.99/98.56\\ 0.63/2.84\end{tabular} &
  \begin{tabular}[c]{@{}c@{}}67.93/99.70\\ 0.53/8.92\end{tabular} &
  \begin{tabular}[c]{@{}c@{}}70.61/77.07\\ 0.62/2.13\end{tabular} \\ \cline{2-9} 
\multicolumn{1}{c|}{} &
  RESISC45 &
  \begin{tabular}[c]{@{}c@{}}94.24/70.29\\ 4.89/0.49\end{tabular} &
  \textit{NA} &
  \begin{tabular}[c]{@{}c@{}}83.27/99.19\\ 3.76/18.44\end{tabular} &
  \begin{tabular}[c]{@{}c@{}}90.56/96.81\\ 2.86/9.63\end{tabular} &
  \begin{tabular}[c]{@{}c@{}}90.46/97.96\\ 4.92/1.23\end{tabular} &
  \begin{tabular}[c]{@{}c@{}}90.62/99.60\\ 2.10/10.10\end{tabular} &
  \begin{tabular}[c]{@{}c@{}}93.21/72.18\\ 2.49/2.18\end{tabular} \\ \cline{2-9} 
\multicolumn{1}{c|}{} &
  EuroSAT &
  \begin{tabular}[c]{@{}c@{}}99.76/72.23\\ 9.26/0.61\end{tabular} &
  \begin{tabular}[c]{@{}c@{}}99.19/83.27\\ 13.54/2.90\end{tabular} &
  \textit{NA} &
  \begin{tabular}[c]{@{}c@{}}98.06/95.90\\ 9.67/19.41\end{tabular} &
  \begin{tabular}[c]{@{}c@{}}98.07/93.97\\ 10.33/2.14\end{tabular} &
  \begin{tabular}[c]{@{}c@{}}96.89/99.46\\ 9.26/9.58\end{tabular} &
  \begin{tabular}[c]{@{}c@{}}99.74/70.64\\ 9.54/1.86\end{tabular} \\ \cline{2-9} 
\multicolumn{1}{c|}{} &
  SVHN &
  \begin{tabular}[c]{@{}c@{}}97.23/67.53\\ 10.53/0.55\end{tabular} &
  \begin{tabular}[c]{@{}c@{}}96.81/90.56\\ 9.16/2.52\end{tabular} &
  \begin{tabular}[c]{@{}c@{}}95.90/98.06\\ 10.16/16.57\end{tabular} &
  \textit{NA} &
  \begin{tabular}[c]{@{}c@{}}94.60/94.26\\ 13.31/2.04\end{tabular} &
  \begin{tabular}[c]{@{}c@{}}92.27/99.38\\ 9.16/8.79\end{tabular} &
  \begin{tabular}[c]{@{}c@{}}97.03/68.09\\ 9.67/3.03\end{tabular} \\ \cline{2-9} 
\multicolumn{1}{c|}{} &
  GTSRB &
  \begin{tabular}[c]{@{}c@{}}98.56/67.99\\ 1.91/0.58\end{tabular} &
  \begin{tabular}[c]{@{}c@{}}97.96/90.46\\ 1.98/2.56\end{tabular} &
  \begin{tabular}[c]{@{}c@{}}93.97/98.07\\ 1.77/15.11\end{tabular} &
  \begin{tabular}[c]{@{}c@{}}94.26/94.60\\ 1.81/9.69\end{tabular} &
  \textit{NA} &
  \begin{tabular}[c]{@{}c@{}}91.58/99.38\\ 1.26/9.82\end{tabular} &
  \begin{tabular}[c]{@{}c@{}}98.20/67.82\\ 1.93/2.66\end{tabular} \\ \cline{2-9} 
\multicolumn{1}{c|}{} &
  MNIST &
  \begin{tabular}[c]{@{}c@{}}99.70/67.92\\ 11.39/0.44\end{tabular} &
  \begin{tabular}[c]{@{}c@{}}99.60/90.62\\ 12.28/3.60\end{tabular} &
  \begin{tabular}[c]{@{}c@{}}99.46/96.89\\ 11.35/9.24\end{tabular} &
  \begin{tabular}[c]{@{}c@{}}99.38/92.27\\ 11.35/18.16\end{tabular} &
  \begin{tabular}[c]{@{}c@{}}99.38/91.58\\ 11.47/2.24\end{tabular} &
  \textit{NA} &
  \begin{tabular}[c]{@{}c@{}}99.70/70.69\\ 11.26/2.29\end{tabular} \\ \cline{2-9} 
\multicolumn{1}{c|}{} &
  DTD &
  \begin{tabular}[c]{@{}c@{}}77.07/70.61\\ 2.13/0.52\end{tabular} &
  \begin{tabular}[c]{@{}c@{}}72.18/93.21\\ 2.07/2.17\end{tabular} &
  \begin{tabular}[c]{@{}c@{}}70.64/99.74\\ 2.23/14.33\end{tabular} &
  \begin{tabular}[c]{@{}c@{}}68.09/97.03\\ 1.97/18.30\end{tabular} &
  \begin{tabular}[c]{@{}c@{}}67.82/98.20\\ 1.91/2.15\end{tabular} &
  \begin{tabular}[c]{@{}c@{}}70.69/99.70\\ 2.07/10.09\end{tabular} &
  \textit{NA} \\ \hline
\end{tabular}%
}
\end{table*}
\textbf{Evaluation TA performance under different scaling coefficient}Since the performance of TA is affected by a merge coefficient, we then measure whether different coefficients affect our proposal. By adjusting the coefficient from 0.3 to 0.8, we observe how the performance of MMP- and MMP+ evolves when combining multiple specialized models, as shown in Figure \ref{fig: scaling coef}. 
Here, the defender is finetuned on EuroSAT, and the free-rider consists of six other tasks. In the figure, we can observe that the performance of MMP- (lines in different shades of blue) is at least larger than 60\%, and the performance of MMP+ (lines in different shades of red) is at most less than 25\%. This demonstrates the effectiveness of our method in preventing free-riders from gaining the specialized capabilities of the defender’s model, regardless of the TA scaling coefficient.

\begin{figure}[t]
    \centering
    \includegraphics[width=1\linewidth]{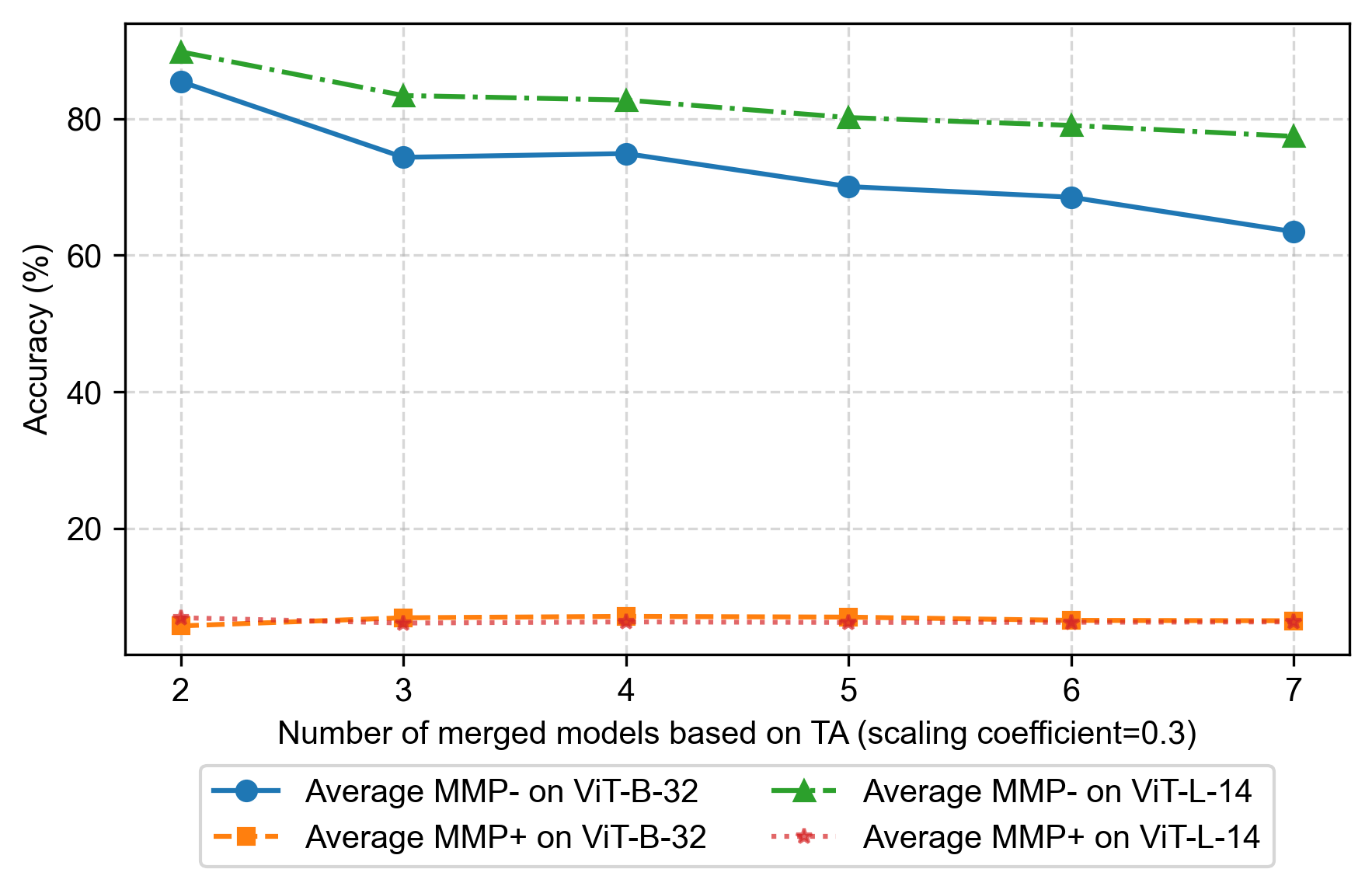}
    \caption{The classification accuracy of MMP-/MMP+ (merged by TA) with different numbers of merged models (2 to 7) Here, defender's model is trained on MNIST, and the other six tasks as free-riders' task) with a scaling coefficient of 0.3, evaluated on ViT-B-32 and ViT-L-14.}
    \label{fig: multiple addition}
\end{figure}

\textbf{Evaluation when more than two models are merged} We further investigate the scenario where more than two models are merged simultaneously with task arithmetic. Specifically, we evaluate the performance of merging 2 to 7 finetuned models at once, examining whether our defense remains effective when merging with more than 2 models. When merging with more than 2 models, the common scaling coefficient is usually set as $\lambda=0.3$, we follow this setting. In Figure \ref{fig: multiple addition}, we can observe performance of MMP- is at least larger than 60\% (green and blue line) while the performance of MMP+ is at most less than 10\% (orange and red line), this means our proposal is still effective with different number of merged model.
The Benign curves (ViT-B-32 in blue, ViT-L-14 in green) show that as the number of merged models increases, the average accuracy gradually declines (e.g., from around 85\% to 63\% on ViT-B-32, and from around 90\% to 77\% on ViT-L-14). In contrast, the Protected curves (ViT-B-32 in orange, ViT-L-14 in red) remain consistently low (around 6–7\%), indicating that once our defense is applied, merging multiple models fails to preserve any meaningful performance. This result demonstrates that our protection method continues to be effective even in scenarios where more than two models are merged.

\textbf{Evaluation on different model merging methods.} Next, we evaluated the effectiveness of our method on other merging approaches. Specifically, we examined four merging methods: TA, TIES, WA, and ADA, along with a plug-in module DARE. Since DARE can be combined with each of the four merging methods to form a new merging approach, we tested a total of eight merging methods. Due to space limitations, we report the average results across all datasets based on each model in Figure \ref{fig: eval different merging}. Full results are shown in the appendix. 

As shown in Figure \ref{fig: eval different merging}, PaRaMS successfully degrades the classification accuracy across all eight merging methods on all three ViT models. Here, the blue bars represent the accuracy of MMP- which merged by TA, TIES, WA, and ADA respectively, The pink bars represent the accuracy of MMP- merged by TA plus DARE, TIES plus DARE, WA plus DARE, and ADA plus DARE, respectively. Similarly, the yellow and green bars indicate the accuracy of MMP+ with or without DARE. A clear gap can be observed between the performance of MMP- and MMP+, demonstrating that our method effectively causes the merging to fail on different merge methods. Furthermore, in this experiment, the accuracy of UMP-/UMP+ (blue and pink dash lines) remains unchanged before and after applying our method. This shows that PaRaMS does not have a negative impact on model performance unless model merging is applied.

Furthermore, We also tested our method with ablation and we have included these results in the appendix.

\begin{figure*}[t]
    \centering
    \includegraphics[width=\linewidth]{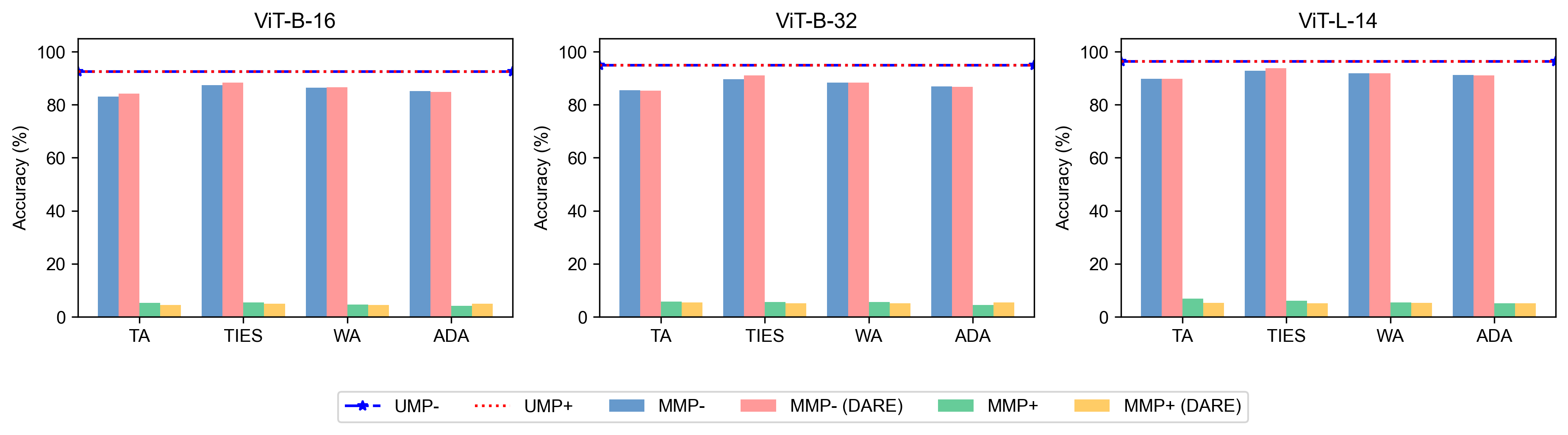}
    \caption{The average classification accuracy of MMP-/MMP+ over all task combinations on three Vision Transformer architectures (ViT-B-16, ViT-B-32, and ViT-L-14). Here, we use different model merging techniques: TA, TIES, WA, and ADA with/without DARE. Dashed and dotted lines indicate baseline performances (UMP and UMP+, respectively).}
    \label{fig: eval different merging}
\end{figure*}

\subsubsection{Results on Image Generation} 

\begin{figure}[t]
    \centering
    \includegraphics[width=1\linewidth]{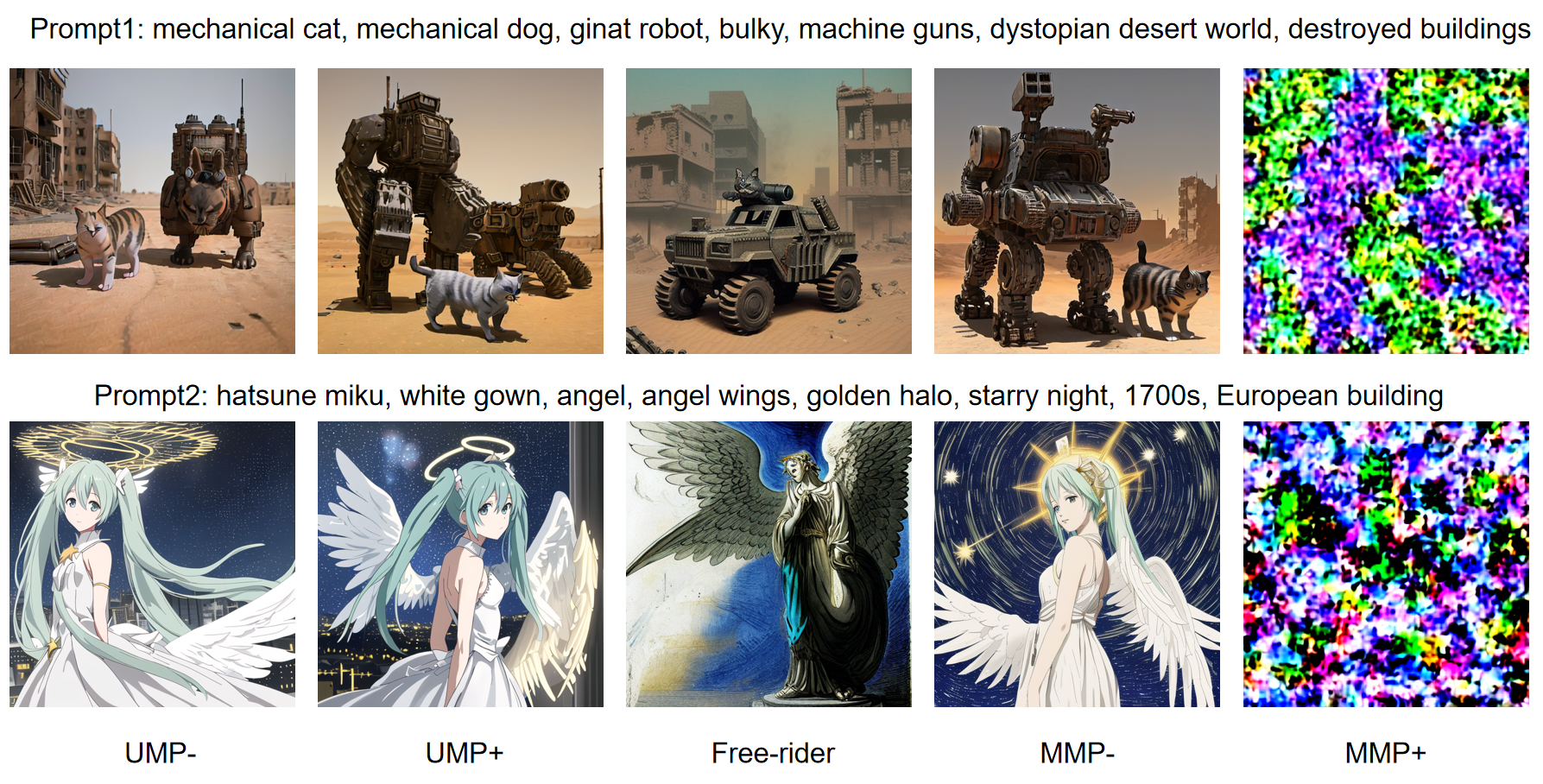}
    \caption{Generated images from UMP-, UMP+, MMP-, MMP+. Each row is related to one prompt set. More examples are shown in the appendix.}
    \label{fig: sd_exp}
\end{figure}

To demonstrate that our proposal is effective in other applications, we evaluated our proposal on a text-to-image generation task. Here, we downloaded finetuned Stable Diffusion 1.5 (SD) \cite{rombach2021highresolution} trained on different image concepts from Hugging Face. By merging these SD by TA, the merged SD is able to generate images containing multiple concepts. For quality assessment, we present the images generated by the model before and after merging and its text-to-image alignment. Following \cite{gu2023mix}, text-to-image alignment was measured by the similarity between the prompt embedding and the generated images using a pretrained CLIP model. A higher similarity indicates better generation quality.

\begin{table}[t]
\centering
\caption{Average text-to-image alignment scores between prompts and 1000 generated images by UMP-, UMP+, free-rider, MMP- and MMP+}
\label{compare_generation_cases}
\begin{tabular}{cccccc}
\hline
      & UMP-   & UMP+   & Free-rider & MMP-   & MMP+   \\ \hline
Prompt1 & 0.3286 & 0.3306 & 0.3245     & 0.3416 & 0.1277 \\ 
Prompt2 & 0.3335 & 0.3428 & 0.2961     & 0.3386 & 0.0820 \\ \hline
\end{tabular}%

\end{table}

In Figure \ref{fig: sd_exp}, we show images generated by UMP-, UMP+, MMP-, and MMP+. We can observe that the images generated UMP+ are clear while images generated by MMP+ (the rightmost images) are almost noise images. Moreover, we synthesize 1000 generated images based on the prompts in the given cases of Figure \ref{fig: sd_exp}, for UMP-, UMP+, MMP-, and MMP+, and measure their average text-to-image similarity. In Table \ref{compare_generation_cases}, we observe that MMP+ has significantly lower text-to-image alignment scores compared to MMP-. And the text-to-image alignment scores of UMP- are closed to UMP+. This means PaRaMS decreases the merged SD model's text-to-image generation ability while keeping the unmerged model's text-to-image generation ability.

\subsubsection{Results on Text Classification}  
We also evaluated our proposal on a different modality to show the wide applicability of PaRaMS other than image domain. In this experiment, we finetuned Llama2 \cite{touvron2023llamaopenefficientfoundation} on the Emotion \cite{emotion} and Twitter datasets \cite{twitter}, which are common text classification tasks, using parameter-efficient finetuning (PEFT) \cite{lora} and full-parameter finetuning. We then applied task arithmetic and PEM-based task arithmetic (PEM-TA) \cite{zhang2023composing} to merge finetuned models with an Llama2 model finetuned on Alpaca. For TA with full parameter finetuning, we protect the defender's model using both parameter scaling and rearrangement. In the PEFT setting, PEM-based TA only merges LoRA-adapted parameters. Since this part of the parameters can be viewed as the multiplication of two matrices, we applied the scaling method only. Table \ref{text_table} shows that the MMP- has high classification performance, while MMP+ performs poorly. For PEM-TA, since we only apply parameter scaling, the performance degradation after merge is relatively small compared to full parameter finetuning setting, but there is still significant performance degradation compared to MMP-. From these results, we can conclude that PaRaMS effectively suppress model merging in the NLP domain, too.

\begin{table}[t]
    \centering
    \caption{The classification accuracy of UMP-/UMP+ and MMP-/MMP+ on two classification task emotion and twitter. Here, the model is Llama2, and the merged methods are TA and PEM-TA}
    \label{text_table}
        \begin{tabular}{lccccc}
            \toprule 
            Dataset & UMP-/UMP+ & MMP-(TA) & MMP+(TA) & MMP-(PEM-TA) & MMP+(PEM-TA)\\ 
            \midrule
            Emotion & 99.8 & 97.6 & 21.4 & 96.1 & 52.7 \\ 
            Twitter & 99.7 & 95.3 & 35.8 & 95.2 & 61.2 \\ 
            \bottomrule
        \end{tabular}
\end{table}

\subsection{Investigation based on Connectivity}
\label{sec_connect}

\begin{figure}[t]
    \centering
    \includegraphics[width=0.9\linewidth]{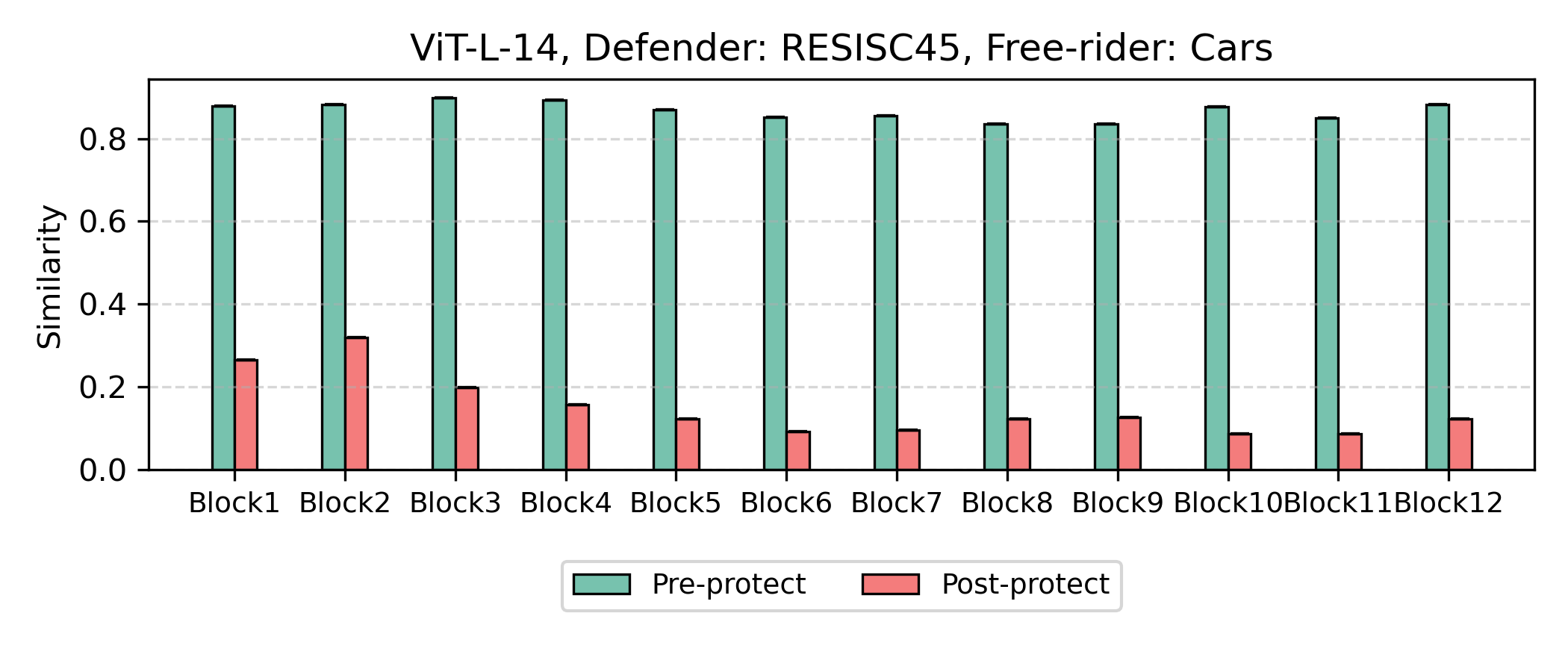}
    \caption{Layer-wise similarity of MMP- and MMP+. The defender model was trained on RESISC45 and a free-rider model trained on Cars. The model architecture is ViT-L-14.}
    \label{fig: similarity_eval}
\end{figure}

As discussed in Sec \ref{sec:insightonMM}, our goal was to force the protected model far away from the basin shared by all finetuned models that trained from the same pretrain model. To verify whether we achieve this goal, we evaluated block-wise similarity between pre/post-protection models and the free-rider's model. The layer-wise similarity evaluates whether interpolating two model parameters with task arithmetic will also result in feature interpolation \cite{zhou2024emergencecrosstasklinearitypretrainingfinetuning}:
$$\text{sim}^{(\ell)}=\text{cosine}[\mathcal{M}_{\text{TA}}^{(\ell)}(\theta_{\text{pre}},\theta_{\text{def}}, \theta_{\text{fr}};x), \frac{1}{2}\theta^{(\ell)}_{\text{def}}(x)+\frac{1}{2}\theta^{(\ell)}_{\text{fr}}(x)].$$
Here $\mathcal{M}_{\text{TA}}^{(\ell)}(\theta_{\text{pre}},\theta_{\text{def}}, \theta_{\text{fr}})$ refers to the output of layer $\ell$ of the merged model (based on TA with a scaling factor $\lambda=0.5$), $\theta^{(\ell)}$ refers to the output of a pre-merging model on layer $\ell$. High similarity indicates two models reside in a shared basin \cite{zhou2024emergencecrosstasklinearitypretrainingfinetuning}. Figure \ref{fig: similarity_eval} shows that laryer-wise similarity of MMP- is much higher than layer-wise similarity of MMP+. This confirms that our objective in Sec.\ref{sec:insightonMM} is achieved.

\section{Possible Contermeasure}

In reality, a ``determined free-rider" may attempt to bypass our proposal by utilizing countermeasures designed against our approach. In this section, we assume free riders who know the technical details of PaRaMS and understand that the published models are protected by PaRaMS. With this assumption, we consider how such an adaptive free-rider would bypass PaRaMS, and then we discuss potential countermeasures and strategies to further defend against them.

\textbf{Adaptive attack.}
First, we consider an adaptive attack to bypass our defense for the MLP part. Recall that the defender’s defense using permutation $\eta_{\text{perm}}$ is designed to maximize the distance between the parameters of the MLP layer in pretrained model and those of the defender’s model. Then, the free-rider might try to recover by finding an adaptive strategy using permutation $\eta'_{\text{perm}}$ that minimizes the distance between the MLP layers in pretrained model and the MLP layers in defender’s model:.
\[
\arg\min_{\eta'_{\text{perm}}} \bigl\|\theta^{\text{MLP}}_{\text{pre}} - \eta'_{\text{perm}} \circ \eta_{\text{perm}}(\theta^{\text{MLP}}_{\mathrm{def}})\bigr\|^2
\]
Here, $\eta'_{\text{perm}}$ refers to the MLP layer permutated by MLP parameter rearrangement. A similar adaptive attack works with the scaling module, too. We denote our protected attention module as $Q'$, $K'$, and $V'$. We assume that free-riders obtain the pretrained model's attention module $Q_{\text{pre}}$, $K_{\text{pre}}$, and $V_{\text{pre}}$. Then, the free-rider can reconvery our scaling strategy by
$\min_{A'}\bigl\| Q_{\text{pre}} - A' Q' \bigr\|^2 $.
Once $A'$ is determined, the scaling factor for $K'$ is set as $A^{'\mathsf{T}}$ immediately. A similar strategy can also be applied to $V'$.

\begin{table}[t]
\centering
\caption{Classification accuracy on merged models under different behavior of free-rider and defender. Here, the models are ViT-B-32 and are merged by TA with coefficient = 0.3. The defender model was fine-tuned on SVHN ($97.46\% \to 96.33\%$ with/without dropout) and free-rider models covering six other tasks.}
\label{adaptive_comp}
\begin{tabular}{ccc}
\hline
Defender       & Free-rider             & Merged models \\ \hline
No Defense     & Non-adaptive           & 84.13\%       \\
PaRaMS         & Non-adaptive           & 7.12\%        \\
PaRaMS+dropout & Non-adaptive           & 7.01\%        \\ \hline
PaRaMS         & Adaptive               & 80.32\%       \\
PaRaMS+dropout & Adaptive               & 64.35\%       \\ \hline
PaRaMS+dropout & knowledge distillation & 82.27\%       \\ \hline
\end{tabular}%
\end{table}

\textbf{Experimental Results under Adaptive Free-riders.}
Table \ref{adaptive_comp} shows the performance of the merged model under different combinations of the behaviors of the defender and free-rider. When the free-rider is not adaptive, the performance of the merged model is significantly degraded (7.12 \%, row 3) irrespective of the defender’s strategy (rows 3-4). However, when the free-rider is adaptive, PaRaMS is bypassed, and the performance of the merged model is almost recovered (80.32\%, row 5). This does not necessarily mean that PaRaMS is meaningless, as a free-rider needs to solve optimization problems and must have prior knowledge of our method. However, we need a defense strategy that can counter adaptive free-riders. 

\textbf{Enhancing Robustness via Dropout-Based Pruning.}
To improve the robustness of PaRaMS and counter against the above adaptive attack, we also propose an additional modification approach that applies dropout-based pruning after using our method. Specifically, for the finetuned model, we set a dropout rate (10\%) and randomly select that fraction of weights to set to zero. While this additional pruning means the model can no longer fully maintain its original functionality, its performance degradation due to dropout is minor empirically (97.46\% to 96.33\% in SVHN). In exchange, the randomness of the dropout makes the recovery by the above adaptive strategy more challenging, gaining resistance against adaptive attacks; the accuracy drop after model merge is almost 20\% (84.13\% to 64.35\% in SVHN, Table \ref{adaptive_comp}, row 6), indicating that even adaptive free-riders cannot restore performance close to the original model. Thus, this minor sacrifice in an unmerged model's accuracy significantly strengthens our defense against adaptive attacks.

\textbf{Evaluating an Ultimate Adaptive Attack.}
Finally, we consider an ultimate adaptive countermeasure. One of the major solutions to transfer a model's ability to other models is knowledge distillation (KD) \cite{Hinton2015DistillingTK}. Since KD does not require the assumption that the models have parameters that are close to each other, a free-rider could leverage models protected with PaRaMS + dropout by using KD. Specifically, they can prepare an unlabeled dataset that is sampled from the same distribution as the protected model's training data and use the protected unmerged model to infer labels. Then, using this labeled dataset, they finetune the pretrained model to obtain a new finetuned model whose parameters have not been perturbed. Finally, the free-rider can merge the newly finetuned model. The last row of Table \ref{adaptive_comp} shows the performance of the merged model through KD. This strategy is expected to bypass all possible proactive defense methods that maintain the performance of UMP. However, the cost is greater than fine-tuning a new model from a pretrained model. Furthermore, it also requires additional data for finetuning. Given these difficulties, reasonable free-riders have little motivation to bypass our method rather than simply train a new model. Hence, we can conclude that this ultimate adaptive attack is impractical. The time cost of KD compared with model merging and our method is proposed in the appendix.

\section{Conclusion}
In this paper, we presented a proactive defense, termed PaRaMS, against model merging, protecting a fine-tuned model’s capabilities from being inherited by free-riders. PaRaMS maintains the pre-merging model's performance while largely reducing the post-merging model's performance. Extensive experiments on image classification, text-to-image generation, and text classification tasks indicate that our proposal is effective on different tasks and model structures. In future work, we plan to investigate owner-controlled merging method, where model owners can specify which models are allowed to be merged, enabling a more secure and customizable merging process.

\bibliography{ref}

\begin{thebibliography}{10}

\bibitem{2018arXiv180308375A}
Abien~Fred {Agarap}.
\newblock {Deep Learning using Rectified Linear Units (ReLU)}.
\newblock {\em arXiv e-prints}, page arXiv:1803.08375, March 2018.

\bibitem{ainsworth2023git}
Samuel Ainsworth, Jonathan Hayase, and Siddhartha Srinivasa.
\newblock Git re-basin: Merging models modulo permutation symmetries.
\newblock In {\em The Eleventh International Conference on Learning Representations}, 2023.

\bibitem{brown2020languagemodelsfewshotlearners}
Tom~B. Brown, Benjamin Mann, Nick Ryder, Melanie Subbiah, Jared Kaplan, Prafulla Dhariwal, Arvind Neelakantan, Pranav Shyam, Girish Sastry, Amanda Askell, Sandhini Agarwal, Ariel Herbert-Voss, Gretchen Krueger, Tom Henighan, Rewon Child, Aditya Ramesh, Daniel~M. Ziegler, Jeffrey Wu, Clemens Winter, Christopher Hesse, Mark Chen, Eric Sigler, Mateusz Litwin, Scott Gray, Benjamin Chess, Jack Clark, Christopher Berner, Sam McCandlish, Alec Radford, Ilya Sutskever, and Dario Amodei.
\newblock Language models are few-shot learners, 2020.

\bibitem{resisc45}
Gong Cheng, Junwei Han, and Xiaoqiang Lu.
\newblock Remote sensing image scene classification: Benchmark and state of the art.
\newblock {\em Proceedings of the IEEE}, 105(10):1865--1883, 2017.

\bibitem{dtddataset}
Mircea Cimpoi, Subhransu Maji, Iasonas Kokkinos, Sammy Mohamed, and Andrea Vedaldi.
\newblock Describing textures in the wild.
\newblock In {\em 2014 IEEE Conference on Computer Vision and Pattern Recognition}, pages 3606--3613, 2014.

\bibitem{Cong2024haveyoumergemymodel}
Tianshuo Cong, Delong Ran, Zesen Liu, Xinlei He, Jinyuan Liu, Yichen Gong, Qi~Li, Anyu Wang, and Xiaoyun Wang.
\newblock Have you merged my model? on the robustness of large language model ip protection methods against model merging.
\newblock In {\em Proceedings of the 1st ACM Workshop on Large AI Systems and Models with Privacy and Safety Analysis}, LAMPS '24, page 69–76, New York, NY, USA, 2024. Association for Computing Machinery.

\bibitem{imagenet2009deng}
Jia Deng, Wei Dong, Richard Socher, Li-Jia Li, Kai Li, and Li~Fei-Fei.
\newblock Imagenet: A large-scale hierarchical image database.
\newblock In {\em 2009 IEEE Conference on Computer Vision and Pattern Recognition}, pages 248--255, 2009.

\bibitem{dosovitskiy2021an}
Alexey Dosovitskiy, Lucas Beyer, Alexander Kolesnikov, Dirk Weissenborn, Xiaohua Zhai, Thomas Unterthiner, Mostafa Dehghani, Matthias Minderer, Georg Heigold, Sylvain Gelly, Jakob Uszkoreit, and Neil Houlsby.
\newblock An image is worth 16x16 words: Transformers for image recognition at scale.
\newblock In {\em International Conference on Learning Representations}, 2021.

\bibitem{gu2023mix}
Yuchao Gu, Xintao Wang, Jay~Zhangjie Wu, Yujun Shi, Yunpeng Chen, Zihan Fan, Wuyou Xiao, Rui Zhao, Shuning Chang, Weijia Wu, et~al.
\newblock Mix-of-show: Decentralized low-rank adaptation for multi-concept customization of diffusion models.
\newblock {\em Advances in Neural Information Processing Systems}, 36:15890--15902, 2023.

\bibitem{resnet2016he}
Kaiming He, Xiangyu Zhang, Shaoqing Ren, and Jian Sun.
\newblock Deep residual learning for image recognition.
\newblock In {\em 2016 IEEE Conference on Computer Vision and Pattern Recognition (CVPR)}, pages 770--778, 2016.

\bibitem{helber2019eurosatnoveldatasetdeep}
Patrick Helber, Benjamin Bischke, Andreas Dengel, and Damian Borth.
\newblock Eurosat: A novel dataset and deep learning benchmark for land use and land cover classification, 2019.

\bibitem{Hinton2015DistillingTK}
Geoffrey~E. Hinton, Oriol Vinyals, and Jeffrey Dean.
\newblock Distilling the knowledge in a neural network.
\newblock {\em ArXiv}, abs/1503.02531, 2015.

\bibitem{lora}
Edward~J Hu, Yelong Shen, Phillip Wallis, Zeyuan Allen-Zhu, Yuanzhi Li, Shean Wang, Lu~Wang, Weizhu Chen, et~al.
\newblock Lora: Low-rank adaptation of large language models.
\newblock {\em ICLR}, 1(2):3, 2022.

\bibitem{ilharco2023editing}
Gabriel Ilharco, Marco~Tulio Ribeiro, Mitchell Wortsman, Ludwig Schmidt, Hannaneh Hajishirzi, and Ali Farhadi.
\newblock Editing models with task arithmetic.
\newblock In {\em The Eleventh International Conference on Learning Representations}, 2023.

\bibitem{carsdataset}
Jonathan Krause, Michael Stark, Jia Deng, and Li~Fei-Fei.
\newblock 3d object representations for fine-grained categorization.
\newblock In {\em 2013 IEEE International Conference on Computer Vision Workshops}, pages 554--561, 2013.

\bibitem{twitter}
Keita Kurita, Paul Michel, and Graham Neubig.
\newblock Weight poisoning attacks on pre-trained models.
\newblock {\em arXiv preprint arXiv:2004.06660}, 2020.

\bibitem{LeCun2005TheMD}
Yann LeCun and Corinna Cortes.
\newblock The mnist database of handwritten digits.
\newblock 2005.

\bibitem{SVHNdataset}
Yuval Netzer, Tao Wang, Adam Coates, Alessandro Bissacco, Bo~Wu, and Andrew~Y. Ng.
\newblock Reading digits in natural images with unsupervised feature learning.
\newblock In {\em NIPS Workshop on Deep Learning and Unsupervised Feature Learning 2011}, 2011.

\bibitem{ortiz-jimenez2023task}
Guillermo Ortiz-Jimenez, Alessandro Favero, and Pascal Frossard.
\newblock Task arithmetic in the tangent space: Improved editing of pre-trained models.
\newblock In {\em Thirty-seventh Conference on Neural Information Processing Systems}, 2023.

\bibitem{radford2021learningtransferablevisualmodels}
Alec Radford, Jong~Wook Kim, Chris Hallacy, Aditya Ramesh, Gabriel Goh, Sandhini Agarwal, Girish Sastry, Amanda Askell, Pamela Mishkin, Jack Clark, Gretchen Krueger, and Ilya Sutskever.
\newblock Learning transferable visual models from natural language supervision, 2021.

\bibitem{rombach2021highresolution}
Robin Rombach, Andreas Blattmann, Dominik Lorenz, Patrick Esser, and Björn Ommer.
\newblock High-resolution image synthesis with latent diffusion models, 2021.

\bibitem{emotion}
Elvis Saravia, Hsien-Chi~Toby Liu, Yen-Hao Huang, Junlin Wu, and Yi-Shin Chen.
\newblock Carer: Contextualized affect representations for emotion recognition.
\newblock In {\em Proceedings of the 2018 conference on empirical methods in natural language processing}, pages 3687--3697, 2018.

\bibitem{gtsrbdataset}
Johannes Stallkamp, Marc Schlipsing, Jan Salmen, and Christian Igel.
\newblock The german traffic sign recognition benchmark: A multi-class classification competition.
\newblock In {\em The 2011 International Joint Conference on Neural Networks}, pages 1453--1460, 2011.

\bibitem{sundar2024multimodal}
Anirudh~S Sundar, Chao-Han~Huck Yang, David~M Chan, Shalini Ghosh, Venkatesh Ravichandran, and Phani~Sankar Nidadavolu.
\newblock Multimodal attention merging for improved speech recognition and audio event classification.
\newblock In {\em 2024 IEEE International Conference on Acoustics, Speech, and Signal Processing Workshops (ICASSPW)}, pages 655--659. IEEE, 2024.

\bibitem{touvron2023llamaopenefficientfoundation}
Hugo Touvron, Thibaut Lavril, Gautier Izacard, Xavier Martinet, Marie-Anne Lachaux, Timothée Lacroix, Baptiste Rozière, Naman Goyal, Eric Hambro, Faisal Azhar, Aurelien Rodriguez, Armand Joulin, Edouard Grave, and Guillaume Lample.
\newblock Llama: Open and efficient foundation language models, 2023.

\bibitem{touvron2023llama}
Hugo Touvron, Louis Martin, Kevin Stone, Peter Albert, Amjad Almahairi, Yasmine Babaei, Nikolay Bashlykov, Soumya Batra, Prajjwal Bhargava, Shruti Bhosale, et~al.
\newblock Llama 2: Open foundation and fine-tuned chat models.
\newblock {\em arXiv preprint arXiv:2307.09288}, 2023.

\bibitem{utans1996weight}
Joachim Utans.
\newblock Weight averaging for neural networks and local resampling schemes.
\newblock In {\em Proc. AAAI-96 Workshop on Integrating Multiple Learned Models. AAAI Press}, pages 133--138. Citeseer, 1996.

\bibitem{pmlr-v162-wortsman22a}
Mitchell Wortsman, Gabriel Ilharco, Samir~Ya Gadre, Rebecca Roelofs, Raphael Gontijo-Lopes, Ari~S Morcos, Hongseok Namkoong, Ali Farhadi, Yair Carmon, Simon Kornblith, and Ludwig Schmidt.
\newblock Model soups: averaging weights of multiple fine-tuned models improves accuracy without increasing inference time.
\newblock In Kamalika Chaudhuri, Stefanie Jegelka, Le~Song, Csaba Szepesvari, Gang Niu, and Sivan Sabato, editors, {\em Proceedings of the 39th International Conference on Machine Learning}, volume 162 of {\em Proceedings of Machine Learning Research}, pages 23965--23998. PMLR, 17--23 Jul 2022.

\bibitem{yadav2023tiesmerging}
Prateek Yadav, Derek Tam, Leshem Choshen, Colin Raffel, and Mohit Bansal.
\newblock {TIES}-merging: Resolving interference when merging models.
\newblock In {\em Thirty-seventh Conference on Neural Information Processing Systems}, 2023.

\bibitem{yamabe2024mergeprintrobustfingerprintingmerging}
Shojiro Yamabe, Tsubasa Takahashi, Futa Waseda, and Koki Wataoka.
\newblock Mergeprint: Robust fingerprinting against merging large language models, 2024.

\bibitem{Survery_ModelMerging_2024}
Enneng Yang, Li~Shen, Guibing Guo, Xingwei Wang, Xiaochun Cao, Jie Zhang, and Dacheng Tao.
\newblock Model merging in llms, mllms, and beyond: Methods, theories, applications and opportunities.
\newblock {\em arXiv preprint arXiv:2408.07666}, 2024.

\bibitem{AdaMerging_ICLR_2024}
Enneng Yang, Zhenyi Wang, Li~Shen, Shiwei Liu, Guibing Guo, Xingwei Wang, and Dacheng Tao.
\newblock Adamerging: Adaptive model merging for multi-task learning.
\newblock {\em The Twelfth International Conference on Learning Representations}, 2024.

\bibitem{yu2024language}
Le~Yu, Bowen Yu, Haiyang Yu, Fei Huang, and Yongbin Li.
\newblock Language models are super mario: Absorbing abilities from homologous models as a free lunch.
\newblock In {\em International Conference on Machine Learning}. PMLR, 2024.

\bibitem{zhang2023composing}
Jinghan Zhang, Junteng Liu, Junxian He, et~al.
\newblock Composing parameter-efficient modules with arithmetic operation.
\newblock {\em Advances in Neural Information Processing Systems}, 36:12589--12610, 2023.

\bibitem{zhou2024emergencecrosstasklinearitypretrainingfinetuning}
Zhanpeng Zhou, Zijun Chen, Yilan Chen, Bo~Zhang, and Junchi Yan.
\newblock On the emergence of cross-task linearity in the pretraining-finetuning paradigm, 2024.

\end{thebibliography}

\newpage
\renewcommand{\thesection}{\Alph{section}}
\appendix

\section{Overview}
In this Appendix, we provide additional details and results that complement the main text:

\begin{itemize}
    \item \textbf{Section} \ref{mm_intro} introduces various model merging methods.
    \item \textbf{Section} \ref{architecture_intro} describes the Transformer-based model architecture used in our experiments, focusing on the MLP and multi-head attention modules.
    \item \textbf{Section} \ref{opt_intro} details the MLP permutation optimization.
    \item \textbf{Section} \ref{equiv_proof} offers a proof of equivalence under our random multi-head scaling, ensuring the model’s functionality remains unchanged.
    \item \textbf{Section} \ref{algo box} presents the complete algorithmic steps of PaRaMS.
    \item \textbf{Section} \ref{datasets} provides dataset descriptions for the image classification tasks.
    \item \textbf{Section} \ref{img_clf} shows additional classification results not included in the main paper, illustrating further evidence of our method’s effectiveness.
    \item \textbf{Section} \ref{img_gen} includes extended experiments on image generation, highlighting the robustness of our defense across diverse prompts and scenarios.
    \item \textbf{Section} \ref{other_res} lists other comparative results between different merging methods.
\end{itemize}

\section{Introductions of Merging Methods}\label{mm_intro}
\paragraph{Weight Average (WA).} WA assumes an equal contribution of each task vector in merged models and merges task vectors of multiple models into a single one by simple averaging: $f(\tau_1,\dots,\tau_n)=\frac{1}{n}\Sigma_{i=1}^{n}\tau_i$.

\paragraph{AdaMerging.} AdaMerging is also based on the weighted sum to aggregate task vectors. Nevertheless, it assumes that the task vector of different layer (i.e, $\tau_i^{\ell}$) has different effects in merging, and proposed a layer-wise coefficients $\Lambda_i=\{\lambda_i^1,\dots, \lambda_i^L\}$. Specifically, merging coefficients $\Lambda_i$ are calculated based on the entropy on an unlabeled held-out dataset, and the merging algorithm is formalized as: $f^{\ell}(\tau_1, \dots, \tau_n)=\lambda_i^{\ell} \Sigma_{i=1}^n\tau_i^{\ell}$ for the $\ell$-th layer. Different from other merging algorithms, additional calculation is required on searching for $\Lambda_i$.

\paragraph{TIES-Merging (TIES).} TIES is a plug-in for model merging methods, which resolve task conflicts in merging by TRIM, ELECT SIGN and MERGE on task vectors. The merging performs as: $f(\tau_1,\dots,\tau_n)=\lambda\Sigma_{i=1}^n\phi(\tau_i)$ where $\phi$ is combined TIES operations and $\lambda=0.3$ maximize the merging performance empirically, the same as TA.

\paragraph{Drop And REscale (DARE).} DARE is also a plug-in for model merging methods like TIES. Following a drop and rescale flow, DARE first perform random drop on $\tau_i$ based on a drop rate $p$ (i.e., setting their values to zeros), and rescales the remaining weights by $1/(1-p)$. DARE often retains or enhances the performance of model merging methods with even 90\% task vectors removed. 

\section{Model Architecture}\label{architecture_intro}
Our method is specially designed for Transformer-based architectures, in which each layer (often called a Transformer block) combines a multi-head attention submodule with a MLP. This design has become a core building block of modern deep learning models, including ViTs, CLIP models, Stable Diffusion models and LLMs such as LLaMA. Since the proposed method is closely related to the structure of MLP and Attention block, we briefly describe how the MLP and Attention submodules operate within each Transformer block. 

\paragraph{MLP.}
The MLP submodule applies nonlinear transformations to each position’s feature vector, often scaling dimensions in the hidden layer. Let $X \in \mathbb{R}^{d}$ be the feature vector at a single position (for simplicity, omitting batch and sequence dimensions). A typical two-layer MLP computes
\[
\mathrm{MLP}(X) \;=\; W_{2} \,\sigma\!\bigl(W_{1}\,X + b_{1}\bigr)\;+\;b_{2},
\]
where $W_1 \in \mathbb{R}^{d_{\mathrm{hidden}} \times d}$, $W_2 \in \mathbb{R}^{d \times d_{\mathrm{hidden}}}$ (with biases $b_{1}, b_{2}$) are learnable parameters, and $\sigma(\cdot)$ is a nonlinear activation (e.g., GELU). This per-position feed-forward step enhances the network’s expressive power without introducing dependencies across positions.

\paragraph{Structure of Multi-head Attention Block.}
Consider having $h$ parallel attention heads, each with dimensionality $d_k$. Suppose the input sequence is represented by
\[
x\in \mathbb{R}^{\mathrm{seq}\times d_{\mathrm{model}}}.
\]
A linear mapping first produces $Q, K, V\in \mathbb{R}^{\mathrm{seq}\times (h\times d_k)}$. We then split these along the last dimension into $h$ parts:
\begin{align*}
&Q \;\to\; \bigl[Q_1,\dots,Q_h\bigr], \\
&K\;\to\; \bigl[K_1,\dots,K_h\bigr], \\
&V \;\to\; \bigl[V_1,\dots,V_h\bigr].
\end{align*} where each $Q_i,K_i,V_i\in\mathbb{R}^{\mathrm{seq}\times d_k}$ corresponds to the $i$-th attention head. For the $i$-th head, the attention output is given by
\[
\mathrm{Attn}\!\bigl(Q_i,K_i,V_i\bigr)
=
\mathrm{softmax}\!\Bigl(\frac{Q_i\,K_i^\mathsf{T}}{\sqrt{d_k}}\Bigr)\,V_i.
\] 
The outputs of the $h$ heads are then concatenated and projected via an output weight $W_O\in \mathbb{R}^{(h\times d_k)\times d_{\mathrm{model}}}$:
\begin{align*}
\mathrm{Attention}\!\bigl(Q,K,V\bigr)
= &\\
\Bigl[
\mathrm{Attn}\!\bigl(Q_1,K_1,V_1\bigr),\,\dots,\,
\mathrm{Attn}\!&\bigl(Q_h,K_h,V_h\bigr)
\Bigr]
\;W_O.
\end{align*}
\section{MLP Permutation Optimization}\label{opt_intro}
Considering the optimization problem in Section Method:
\[
\arg\max_{\eta_{\text{perm}}} \bigl\|\theta^{\text{MLP}}_{\text{pre}} - \eta_{\text{perm}}(\theta^{\text{MLP}}_{\text{def}})\bigr\|^2=\arg\min_{\eta_{\text{perm}}} \, \theta^{\text{MLP}}_{\text{pre}} \cdot \eta_{\text{perm}}(\theta^{\text{MLP}}_{\text{def}}).
\]
Which can be re-expressed in the following term in a 2-layer MLP:
\begin{align*}
\arg\min_{\eta_{\text{perm}}=\{P_i\}}\sum_{i=1}^{n}[\langle W^{(i)}_{\text{premlp1}}, P_iW^{(i)}_{\text{defmlp1}}  \rangle_F \\
+\langle W^{(i)}_{\text{premlp2}}, W^{(i)}_{\text{defmlp2}}P_i^{\mathsf{T}} \rangle_F].
\end{align*}
where $\langle A, B\rangle_F$ denotes the Frobenius inner productbetween real-valued matrices $A$ and $B$. Hence, the optimization could be re-expressed and solved as a linear assignment problem.

\section{Proof of Equivelance based on Random Scaling}\label{equiv_proof}
First, scaling on \(Q_i\) and \(K_i\) keeps attention weights unchanged. Suppose we multiply \(Q_i\) by a diagonal matrix \(A_i\) and simultaneously multiply \(K_i\) by \(A^{-1}_i\). Then
\[
\frac{Q_i'\,K_i'^\mathsf{T}}{\sqrt{d_k}}
    \;=\;
\frac{(A_i\,Q_i)\,\bigl(A_i^{-1}\,K_i\bigr)^\mathsf{T}}{\sqrt{d_k}}
\;=\;
\frac{Q_i\,K_i^\mathsf{T}}{\sqrt{d_k}},
\]
ensuring that the attention score matrix and thus the softmax weights remain identical to the original.
  
Scaling \(V_i\) and the output projection \(W_O\) is also an inverse pair.
We could multiply \(V_i\) by diagonal matrix \(B_i\) (possibly channelwise or headwise) and compensate by multiplying the corresponding block in the output projection by \(B^{-1}\). Concretely, the single-head output keeps identical
\begin{align*}    
&\mathrm{Attention}\bigl(Q_i',K_i',V_i'\bigr)
= \\ \mathrm{softmax}\!&\Bigl(\frac{Q_i\,K_i^\mathsf{T}}{\sqrt{d_k}}\Bigr)\,\bigl(B_iV_i\bigr)W_O[:, i]B_i^{-1}=\\
&\mathrm{Attention}\bigl(Q_i,K_i,V_i\bigr),
\end{align*}
Hence, each head \(i\) can adopt
\[
  Q_i' = A_i \,Q_i,\,\,
  K_i' = A_i^{-1}\,K_i,\,\,
  V_i' = B_i\,V_i,\,\,
  W_O'[:,i] = W_O[:,i]\,B_i^{-1}.
\]
so that the multi-head attention output ends up identical to its original form.

\section{Algorithm of PaRaMS}\label{algo box}
The algorithm consists of 2 subblocks: Parameter Rearrangement for MLP block and Random Multi-head Scaling for attention block. The pseudocode for our method is shown in Algorithm \ref{alg:params}.
\begin{algorithm}[htbp]
\caption{PaRaMS}
\label{alg:params}
\textbf{Input:} Fine-tuned model $\theta_{\mathrm{def}}$, pretrained checkpoint $\theta_{\mathrm{pre}}$, scaling range $[s_{\min}, s_{\max}]$\\
\textbf{Output:} Modified model $\hat{\theta}$

\begin{algorithmic}[1]
\State $\hat{\theta} \gets \theta_{\mathrm{def}}$
\vspace{4pt}
\Statex \textbf{Step 1: MLP Parameter Rearrangement}
\For{each MLP layer $\ell$}
   \State $P^{(\ell)} \;\gets\; \displaystyle\arg\max_{P}\;\Bigl\|\;P^{(\ell)}\bigl(\theta_{\mathrm{def}}^{(\ell)}\bigr)\;-\;\theta_{\mathrm{pre}}^{(\ell)}\Bigr\|^2$
   \State $W_{1}^{(\ell)} \;\gets\; P^{(\ell)}\,W_{1}^{(\ell)}$
   \State $W_{2}^{(\ell)} \;\gets\; W_{2}^{(\ell)}\,\bigl(P^{(\ell)}\bigr)^\top$
   \State $b_{1}^{(\ell)} \;\gets\; P^{(\ell)}\,b_{1}^{(\ell)}$
\EndFor
\vspace{4pt}
\Statex \textbf{Step 2: Random Multi-Head Scaling}
\For{each layer $j$ containing multi-head attention}
   \For{each attention head $i$}
       \State $\mathbf{a}_i \sim \mathcal{U}(s_{\min}, s_{\max})^{d_k}$
       \State $A_i \gets \mathrm{diag}(\mathbf{a}_i)\in\mathbb{R}^{d_k\times d_k}$
       \State $Q_{j,i} \;\gets\; A_i\,Q_{j,i}$
       \State $K_{j,i} \;\gets\; A_i^{-1}K_{j,i}$
       \vspace{2pt}
       \State $\mathbf{b}_i \sim \mathcal{U}(s_{\min}, s_{\max})^{d_k}$
       \State $B_i \gets \mathrm{diag}(\mathbf{b}_i)\in\mathbb{R}^{d_k\times d_k}$
       \State $V_{j,i} \;\gets\; B_i\,V_{j,i}$
       \State $W_{\mathrm{out}}[j,:,i] \;\gets\; W_{\mathrm{out}}[j,:,i] \cdot B_i^{-1}$
   \EndFor
\EndFor
\vspace{2pt}
\State \textbf{return} $\hat{\theta}$
\end{algorithmic}
\end{algorithm}

\section{Dataset Discriptions used in Image Classification Task}\label{datasets}
\paragraph{Cars.} 
Cars dataset comprises high-resolution images of cars, with a focus on fine-grained vehicle classification. It contains about 16,000 images spanning 196 car models, making it a benchmark for fine-grained recognition tasks.

\paragraph{RESISC45.}
RESISC45 is a remote sensing image dataset containing 31,500 images from 45 scene classes (e.g., airports, industrial areas, harbors). Each class has 700 images, facilitating the study of aerial scene classification.

\paragraph{SVHN.}
SVHN dataset features real-world digit images extracted from Google Street View. It includes over 600,000 labeled digits, commonly used for digit recognition under challenging, cluttered backgrounds.

\paragraph{GTSRB.}
GTRSB consists of over 50,000 images covering 43 traffic sign classes. It is widely used to evaluate classification performance in real-world traffic scenarios.

\paragraph{MNIST.}
MNIST is a classic dataset of handwritten digit images (0–9), comprising 70,000 grayscale images (60,000 for training, 10,000 for testing). It remains a foundational benchmark for evaluating basic image classification methods.

\paragraph{EuroSAT.}
EuroSAT is a satellite image dataset derived from Sentinel-2 data, covering 10 land-use and land-cover classes (e.g., forest, residential). It contains 27,000 labeled images, serving as a testbed for remote-sensing scene classification.

\paragraph{DTD.}
DTD Dataset includes 5,640 images of texture patterns grouped into 47 classes (e.g., banded, porous, grid). It focuses on texture-centric classification and is often used to assess a model’s ability to capture fine-grained visual attributes.

\section{Other Results on Image Classification}\label{img_clf}
\subsection{Ablation Study}
\label{sec:ablation}
Our method comprises two key modules, MLP parameter rearrangement and random multi-head scaling. They jointly disrupt model merging while preserving the model’s functionality. To examine how each module individually contributes to this disruption, we perform an ablation study by omitting one module at a time. Specifically, we compare the avarage accuracy of MMP+ based on ViT-L-14 and TA (scaling coefficient=0.8). Results are shown in Table \ref{ablation}.

\begin{table}[t]
\centering
\caption{Ablation study based on ViT-L-14. Defender: Cars. Free-rider: other six tasks. Merging method: TA (scaling coefficient=0.8).}
\label{ablation}
\begin{tabular}{cc}
\hline
Setting            & Average accuracy (\%) of MMP+ \\ \hline
Rearrangement Only & 5.90                          \\ \hline
Scaling Only       & 10.92                         \\ \hline
PaRaMS             & 4.99                          \\ \hline
\end{tabular}%

\end{table}

Table \ref{ablation} shows that even with a single module (Rearrangement Only or Scaling Only), the MMP+ accuracy still exhibits significant degradation compared to benign merging scenarios. This indicates that either MLP rearrangement or random multi-head scaling alone can effectively disrupt model merging, thereby providing flexibility when only one type of parameter manipulation is applicable in merging. Nonetheless, employing both modules (PaRaMS) yields an even stronger defense, as it further increases the parameter distance across both MLP and attention modules. We thus conclude that each module individually contributes substantially to disrupting model merging, yet their combination provides a more robust protection.

\subsection{Computation Cost Comparison} We further compare the computational costs associated with the original model merging, our proposed defense (PaRaMS), and an ultimate adaptive attack based on knowledge distillation (KD). This experiment is conducted on a Windows 11 platform equipped with AMD Ryzen 9 9950X CPU and Nvidia A6000 Ada GPU. For the KD setting, we fine-tune the model for 50 epochs anda a 128 batch size. The average computation time across seven datasets is presented in Table \ref{comp_cost}.

\begin{table}[t]
\centering
\caption{Average computational cost  comparison of merging based on TA, applying PaRaMS and performing knowledge distillation as adaptive method. }
\label{comp_cost}
\begin{tabular}{ccc}
\hline
Setting/Model          & ViT-B-32 & ViT-L-14 \\ \hline
Task Arithmetic        & 1.59s     & 5.08s     \\ \hline
PaRaMS                 & 57.32s    & 173.05s   \\ \hline
Knowledge Distillation &    3.82h     &    11.42h     \\ \hline
Finetune From Pretrain &       2.57h       &    7.83h       \\ \hline
\end{tabular}%

\end{table}

As shown in Table \ref{comp_cost}, PaRaMS is computationally lightweight, requiring only 57.32 seconds on average for ViT-B-32 and 173.05 seconds for ViT-L-14. In comparison, the standard model merging based on Task Arithmetic is extremely efficient, taking less than 10 seconds, thus significantly lowering the barrier for potential free-riders, and highlighting the risks of intellectual property (IP) infringement associated with lightweight merging methods. On the other hand, the adaptive attack via KD incurs substantially higher computational costs (near 4 hours for ViT-B-32 and 12 hours for ViT-L-14). Moreover, it requires access to the defender’s private training data, which is typically unavailable since model publishers rarely share their proprietary datasets publicly. Therefore, the KD approach is not practically feasible in most open-source scenarios.

\section{Other Results on Image Generation}\label{img_gen}
\begin{figure}
    \centering
    \includegraphics[width=1\linewidth]{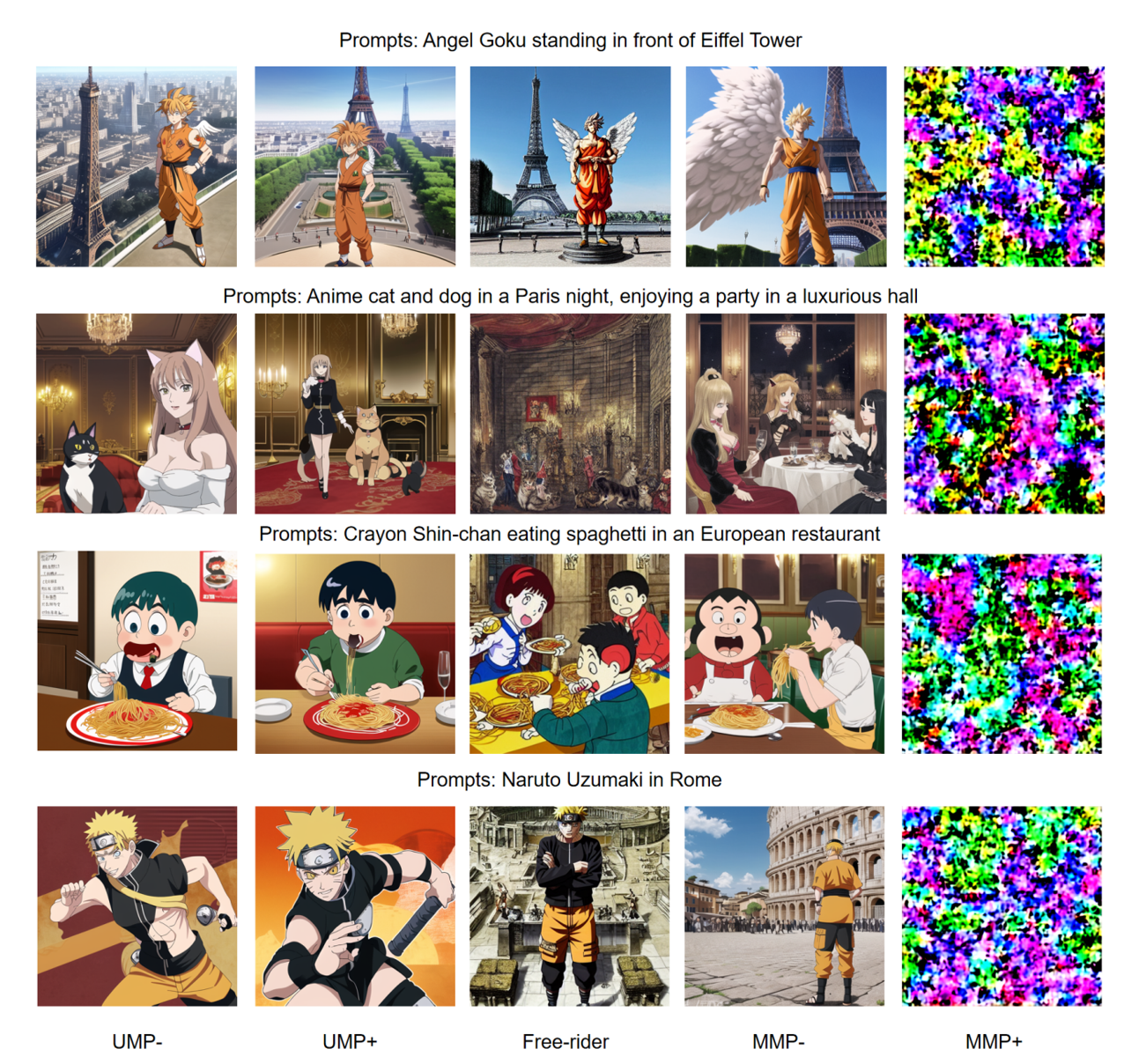}
    \caption{Generated images from UMP-, UMP+, MMP-, MMP+. Each row is related to one prompt set.}
    \label{sd_examples_1}
\end{figure}

We further show several sets of generated images from UMP-, UMP+, free-rider, MMP- and MMP+. For Figure \ref{sd_examples_1}, the defender is an Anime-based SD1.5 model, and the free-rider is a reality-Europe SD1.5 model, both from HuggingFace. 

In these generated images, UMP- (the unprotected single model) and UMP+ (the protected single model) both faithfully capture the prompt details—such as “Angel Goku” standing before the Eiffel Tower or “Naruto Uzumaki” in Rome—indicating that our defense does not degrade the protected model’s original generative performance. Conversely, MMP- (the benign merged model) successfully blends concepts from both the defender and free-rider, as shown by coherent scenes like “Crayon Shin-chan eating spaghetti in an European restaurant.”

Once protection is applied, however, MMP+ (the protected merged model) fails to inherit the defender’s specialized knowledge, resulting in heavily distorted or random artifacts. For instance, the final column in each row often appears corrupted or nonsensical, demonstrating that the free-rider cannot exploit the protected model’s fine-tuned capabilities through merging. Overall, these results underscore that our defense preserves the defender’s performance while rendering merged outputs unusable.

\begin{figure}
    \centering
    \includegraphics[width=1\linewidth]{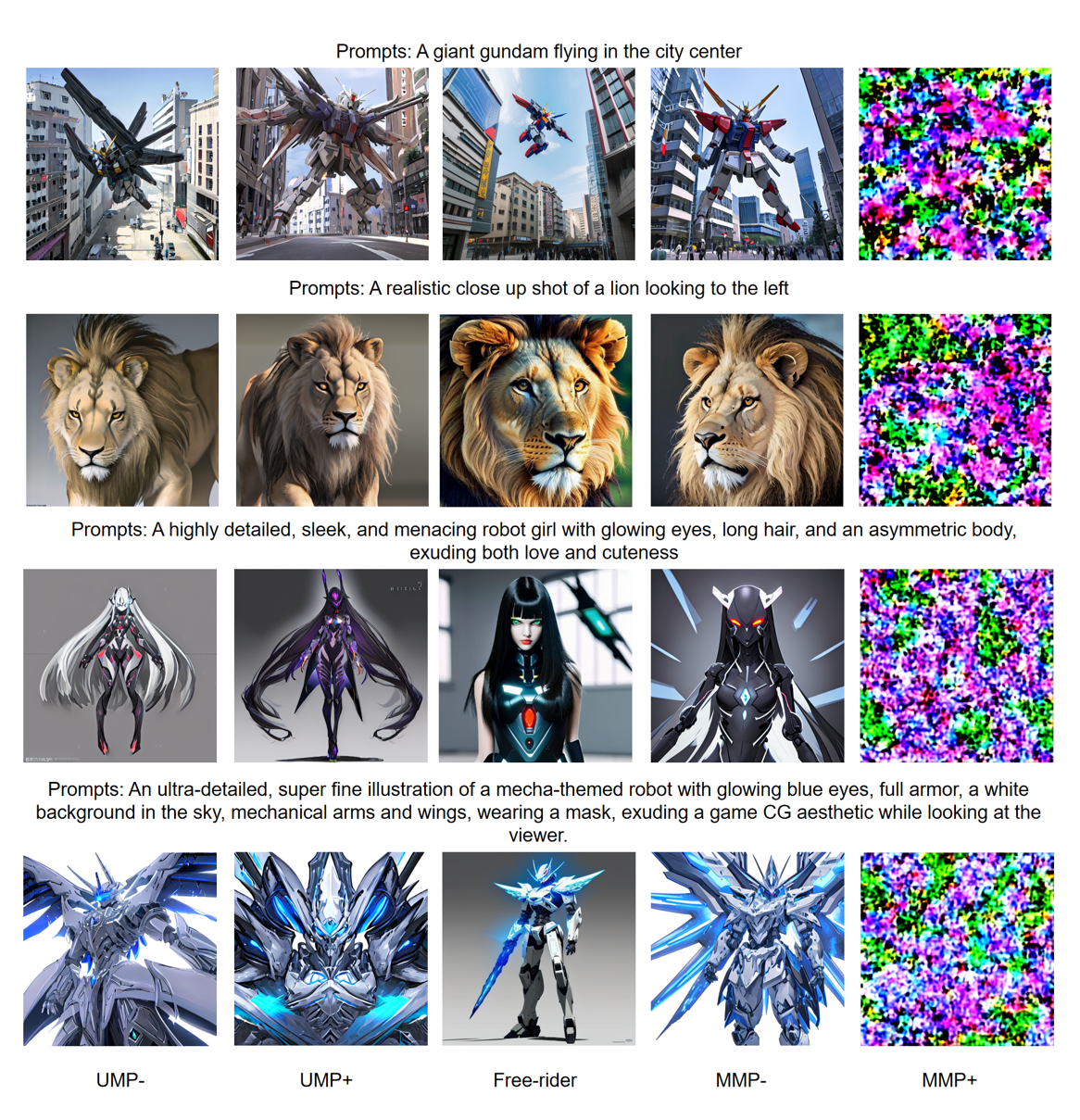}
    \caption{Generated images from UMP-, UMP+, MMP-, MMP+. Each row is related to one prompt set.}
    \label{sd_examples_2}
\end{figure}

We also show generation examples of Animated-Gundam style and realistic style in  Figure \ref{sd_examples_2}.

In these examples, UMP-/UMP+ specializes in an anime–mecha aesthetic (e.g., “giant gundam” and “robot girl”), while the free-rider focuses on more realistic imagery (e.g., “close-up shot of a lion”). The UMP- and UMP+ both render their respective prompts with high fidelity—indicating our defense does not degrade the defender’s own mecha–anime style. Meanwhile, the free-rider model excels at realism, as seen in the lion images.

For MMP-, the outputs successfully blend the anime mecha concepts with the free-rider’s realistic style, producing coherent “hybrid” results. However, once the defender is protected, MMP+ fails to inherit the mecha–anime capabilities and instead generates heavily corrupted or noisy outputs. This underscores that while our defense preserves the defender’s fine-tuned strengths, it prevents any merged model from cheaply acquiring those specialized skills.

\section{Comprehensive Results of Different Merging Methods}\label{other_res}
We show one set of results on TA without DARE based on Vit-B-32 in the paper, and overall average accuracy of eight merging settings on three architectures. Here we show the other seven settings of ViT-B-32 in the following. 

The tables show a same result that our method efficiently disturb all the following settings on all datasets.

\begin{table*}[t]
\centering
\caption{The classification accuracy of MMP-/MMP+ (merged by TA with DARE) on $\mathcal{T}_{\text{def}}$ and $\mathcal{T}_{\text{fr}}$ on ViT-B-32 ($\lambda=0.8$).}
\label{tadare_vitb32_cpr}
\resizebox{\textwidth}{!}{%
\begin{tabular}{cc|ccccccc}
\hline
\multicolumn{2}{l|}{\multirow{2}{*}{\begin{tabular}[c]{@{}l@{}}MMP- Accuracy (\%) on $\mathcal{T}_{\text{def}}$/$\mathcal{T}_{\text{fr}}$\\ MMP+ Accuracy (\%) on $\mathcal{T}_{\text{def}}$/$\mathcal{T}_{\text{fr}}$\end{tabular}}} &
  \multicolumn{7}{c}{$\mathcal{T}_{\text{fr}}$} \\ \cline{3-9} 
\multicolumn{2}{l|}{} &
  Cars &
  RESISC45 &
  EuroSAT &
  SVHN &
  GTSRB &
  MNIST &
  DTD \\ \hline
\multicolumn{1}{c|}{\multirow{7}{*}{$\mathcal{T}_{\text{def}}$}} &
  Cars &
  \textit{NA} &
  \begin{tabular}[c]{@{}c@{}}69.93/93.95\\ 0.52/2.84\end{tabular} &
  \begin{tabular}[c]{@{}c@{}}71.99/99.80\\ 0.55/9.93\end{tabular} &
  \begin{tabular}[c]{@{}c@{}}67.65/97.30\\ 0.60/8.36\end{tabular} &
  \begin{tabular}[c]{@{}c@{}}67.53/98.53\\ 0.47/3.35\end{tabular} &
  \begin{tabular}[c]{@{}c@{}}67.62/99.67\\ 0.52/9.00\end{tabular} &
  \begin{tabular}[c]{@{}c@{}}70.17/76.70\\ 0.39/1.65\end{tabular} \\ \cline{2-9} 
\multicolumn{1}{c|}{} &
  RESISC45 &
  \begin{tabular}[c]{@{}c@{}}93.95/69.93\\ 2.33/0.56\end{tabular} &
  \textit{NA} &
  \begin{tabular}[c]{@{}c@{}}83.14/99.90\\ 2.60/3.83\end{tabular} &
  \begin{tabular}[c]{@{}c@{}}90.51/96.82\\ 1.95/7.41\end{tabular} &
  \begin{tabular}[c]{@{}c@{}}90.02/97.86\\ 1.76/0.82\end{tabular} &
  \begin{tabular}[c]{@{}c@{}}90.49/99.57\\ 1.22/11.16\end{tabular} &
  \begin{tabular}[c]{@{}c@{}}93.22/72.45\\ 2.11/2.13\end{tabular} \\ \cline{2-9} 
\multicolumn{1}{c|}{} &
  EuroSAT &
  \begin{tabular}[c]{@{}c@{}}99.80/71.99\\ 9.15/0.53\end{tabular} &
  \begin{tabular}[c]{@{}c@{}}99.90/83.14\\ 9.33/2.32\end{tabular} &
  \textit{NA} &
  \begin{tabular}[c]{@{}c@{}}98.11/95.89\\ 11.13/9.00\end{tabular} &
  \begin{tabular}[c]{@{}c@{}}98.11/93.33\\ 10.65/3.80\end{tabular} &
  \begin{tabular}[c]{@{}c@{}}96.98/99.48\\ 19.04/9.66\end{tabular} &
  \begin{tabular}[c]{@{}c@{}}99.72/70.48\\ 12.44/1.81\end{tabular} \\ \cline{2-9} 
\multicolumn{1}{c|}{} &
  SVHN &
  \begin{tabular}[c]{@{}c@{}}97.30/67.65\\ 9.57/0.47\end{tabular} &
  \begin{tabular}[c]{@{}c@{}}96.82/90.51\\ 9.23/1.79\end{tabular} &
  \begin{tabular}[c]{@{}c@{}}95.89/98.11\\ 11.46/8.78\end{tabular} &
  \textit{NA} &
  \begin{tabular}[c]{@{}c@{}}94.64/94.18\\ 8.53/2.22\end{tabular} &
  \begin{tabular}[c]{@{}c@{}}92.27/99.40\\ 9.96/10.88\end{tabular} &
  \begin{tabular}[c]{@{}c@{}}96.95/67.77\\ 9.21/2.34\end{tabular} \\ \cline{2-9} 
\multicolumn{1}{c|}{} &
  GTSRB &
  \begin{tabular}[c]{@{}c@{}}98.53/67.53\\ 1/59/0.62\end{tabular} &
  \begin{tabular}[c]{@{}c@{}}97.86/90.02\\ 1.47/2.57\end{tabular} &
  \begin{tabular}[c]{@{}c@{}}93.33/98.11\\ 1.85/9.31\end{tabular} &
  94.18/96.64 &
  \textit{NA} &
  \begin{tabular}[c]{@{}c@{}}91.96/99.35\\ 2.03/9.24\end{tabular} &
  \begin{tabular}[c]{@{}c@{}}98.18/67.39\\ 2.41/2.61\end{tabular} \\ \cline{2-9} 
\multicolumn{1}{c|}{} &
  MNIST &
  \begin{tabular}[c]{@{}c@{}}99.67/67.62\\ 9.71/0.47\end{tabular} &
  \begin{tabular}[c]{@{}c@{}}99.57/90.49\\ 9.51/2.19\end{tabular} &
  \begin{tabular}[c]{@{}c@{}}99.48/96.98\\  8.54/11.50\end{tabular} &
  \begin{tabular}[c]{@{}c@{}}99.40/92.27\\ 9.82/9.07\end{tabular} &
  \begin{tabular}[c]{@{}c@{}}99.35/91.96\\ 9.72/3.28\end{tabular} &
  \textit{NA} &
  \begin{tabular}[c]{@{}c@{}}99.67/70.21\\ 10.36/2.18\end{tabular} \\ \cline{2-9} 
\multicolumn{1}{c|}{} &
  DTD &
  \begin{tabular}[c]{@{}c@{}}76.70/70.17\\ 2.55/0.45\end{tabular} &
  \begin{tabular}[c]{@{}c@{}}72.45/93.22\\ 2.23/1.73\end{tabular} &
  \begin{tabular}[c]{@{}c@{}}70.48/99.72\\ 1.76/11.93\end{tabular} &
  \begin{tabular}[c]{@{}c@{}}67.77/96.95\\ 2.18/8.02\end{tabular} &
  \begin{tabular}[c]{@{}c@{}}67.39/98.18\\ 2.07/1.69\end{tabular} &
  \begin{tabular}[c]{@{}c@{}}70.21/99.67\\ 2.18/9.94\end{tabular} &
  \textit{NA} \\ \hline
\end{tabular}%
}
\end{table*}

\begin{table*}[t]
\centering
\caption{The classification accuracy of MMP-/MMP+ (merged by SA) on $\mathcal{T}_{\text{def}}$ and $\mathcal{T}_{\text{fr}}$ on ViT-B-32.}
\label{sa_vitb32_cpr}
\resizebox{\textwidth}{!}{%
\begin{tabular}{cc|ccccccc}
\hline
\multicolumn{2}{l|}{\multirow{2}{*}{\begin{tabular}[c]{@{}l@{}}MMP- Accuracy (\%) on $\mathcal{T}_{\text{def}}$/$\mathcal{T}_{\text{fr}}$\\ MMP+ Accuracy (\%) on $\mathcal{T}_{\text{def}}$/$\mathcal{T}_{\text{fr}}$\end{tabular}}} &
  \multicolumn{7}{c}{$\mathcal{T}_{\text{fr}}$} \\ \cline{3-9} 
\multicolumn{2}{l|}{} &
  Cars &
  RESISC45 &
  EuroSAT &
  SVHN &
  GTSRB &
  MNIST &
  DTD \\ \hline
\multicolumn{1}{c|}{\multirow{7}{*}{$\mathcal{T}_{\text{def}}$}} &
  Cars &
  \textit{NA} &
  \begin{tabular}[c]{@{}c@{}}73.20/92.87\\ 0.46/2.68\end{tabular} &
  \begin{tabular}[c]{@{}c@{}}73.60/99.30\\ 0.46/9.24\end{tabular} &
  \begin{tabular}[c]{@{}c@{}}71.87/95.36\\ 0.46/7.56\end{tabular} &
  \begin{tabular}[c]{@{}c@{}}73.42/96.89\\ 0.46/3.09\end{tabular} &
  \begin{tabular}[c]{@{}c@{}}73.00/99.41\\ 0.46/9.58\end{tabular} &
  \begin{tabular}[c]{@{}c@{}}74.28/72.07\\ 0.46/2.13\end{tabular} \\ \cline{2-9} 
\multicolumn{1}{c|}{} &
  RESISC45 &
  \begin{tabular}[c]{@{}c@{}}92.87/73.20\\ 2.62/0.51\end{tabular} &
  \textit{NA} &
  \begin{tabular}[c]{@{}c@{}}87.60/98.93\\ 2.27/10.69\end{tabular} &
  \begin{tabular}[c]{@{}c@{}}90.92/94.66\\ 2.94/8.11\end{tabular} &
  \begin{tabular}[c]{@{}c@{}}91.11/95.33\\ 3.05/1.73\end{tabular} &
  \begin{tabular}[c]{@{}c@{}}91.10/99.43\\ 3.73/9.24\end{tabular} &
  \begin{tabular}[c]{@{}c@{}}92.14/70.11\\ 3.06/2.39\end{tabular} \\ \cline{2-9} 
\multicolumn{1}{c|}{} &
  EuroSAT &
  \begin{tabular}[c]{@{}c@{}}99.30/73.60\\ 9.17/0.52\end{tabular} &
  \begin{tabular}[c]{@{}c@{}}98.93/87.60\\ 12.93/2.52\end{tabular} &
  \textit{NA} &
  \begin{tabular}[c]{@{}c@{}}97.67/92.97\\ 13.85/13.79\end{tabular} &
  \begin{tabular}[c]{@{}c@{}}97.11/92.24\\ 10.94/2.14\end{tabular} &
  \begin{tabular}[c]{@{}c@{}}97.96/99.18\\ 8.52/9.84\end{tabular} &
  \begin{tabular}[c]{@{}c@{}}99.15/69.31\\ 13.00/2.13\end{tabular} \\ \cline{2-9} 
\multicolumn{1}{c|}{} &
  SVHN &
  \begin{tabular}[c]{@{}c@{}}95.36/71.87\\ 9.69/0.36\end{tabular} &
  \begin{tabular}[c]{@{}c@{}}94.66/90.92\\ 9.69/2.57\end{tabular} &
  \begin{tabular}[c]{@{}c@{}}92.97/97.67\\ 9.69/9.67\end{tabular} &
  \textit{NA} &
  \begin{tabular}[c]{@{}c@{}}94.74/94.83\\ 9.69/3.16\end{tabular} &
  \begin{tabular}[c]{@{}c@{}}92.50/99.37\\ 9.69/9.82\end{tabular} &
  \begin{tabular}[c]{@{}c@{}}95.36/70.53\\ 9.69/2.82\end{tabular} \\ \cline{2-9} 
\multicolumn{1}{c|}{} &
  GTSRB &
  \begin{tabular}[c]{@{}c@{}}96.89/73.42\\ 3.09/0.44\end{tabular} &
  \begin{tabular}[c]{@{}c@{}}95.33/91.11\\ 3.09/2.52\end{tabular} &
  \begin{tabular}[c]{@{}c@{}}92.24/97.11\\ 3.18/14.54\end{tabular} &
  \begin{tabular}[c]{@{}c@{}}94.83/94.74\\ 3.09/8.19\end{tabular} &
  \textit{NA} &
  \begin{tabular}[c]{@{}c@{}}93.25/99.12\\ 3.17/8.95\end{tabular} &
  \begin{tabular}[c]{@{}c@{}}96.78/70.85\\ 3.09/2.13\end{tabular} \\ \cline{2-9} 
\multicolumn{1}{c|}{} &
  MNIST &
  \begin{tabular}[c]{@{}c@{}}99.41/73.00\\ 9.26/0.47\end{tabular} &
  \begin{tabular}[c]{@{}c@{}}99.43/91.10\\ 8.61/2.10\end{tabular} &
  \begin{tabular}[c]{@{}c@{}}99.18/97.96\\ 9.78/13.80\end{tabular} &
  \begin{tabular}[c]{@{}c@{}}99.37/92.50\\ 9.82/9.70\end{tabular} &
  \begin{tabular}[c]{@{}c@{}}99.12/93.25\\ 9.82/5.87\end{tabular} &
  \textit{NA} &
  \begin{tabular}[c]{@{}c@{}}99.45/69.10\\ 9.82/2.29\end{tabular} \\ \cline{2-9} 
\multicolumn{1}{c|}{} &
  DTD &
  \begin{tabular}[c]{@{}c@{}}72.07/74.28\\ 1.81/0.62\end{tabular} &
  \begin{tabular}[c]{@{}c@{}}70.11/92.14\\ 1.76/2.44\end{tabular} &
  \begin{tabular}[c]{@{}c@{}}69.31/99.15\\ 1.76/19.67\end{tabular} &
  \begin{tabular}[c]{@{}c@{}}70.53/95.36\\ 1.65/14.74\end{tabular} &
  \begin{tabular}[c]{@{}c@{}}70.85/96.78\\ 1.86/2.91\end{tabular} &
  \begin{tabular}[c]{@{}c@{}}69.10/99.45\\ 1.91/11.02\end{tabular} &
  \textit{NA} \\ \hline
\end{tabular}%
}
\end{table*}

\begin{table*}[t]
\centering
\caption{The classification accuracy of MMP-/MMP+ (merged by SA with DARE) on $\mathcal{T}_{\text{def}}$ and $\mathcal{T}_{\text{fr}}$ on ViT-B-32.}
\label{Sadare_vitb32_cpr}
\resizebox{\textwidth}{!}{%
\begin{tabular}{cc|ccccccc}
\hline
\multicolumn{2}{l|}{\multirow{2}{*}{\begin{tabular}[c]{@{}l@{}}MMP- Accuracy (\%) on $\mathcal{T}_{\text{def}}$/$\mathcal{T}_{\text{fr}}$\\ MMP+ Accuracy (\%) on $\mathcal{T}_{\text{def}}$/$\mathcal{T}_{\text{fr}}$\end{tabular}}} &
  \multicolumn{7}{c}{$\mathcal{T}_{\text{fr}}$} \\ \cline{3-9} 
\multicolumn{2}{l|}{} &
  Cars &
  RESISC45 &
  EuroSAT &
  SVHN &
  GTSRB &
  MNIST &
  DTD \\ \hline
\multicolumn{1}{c|}{\multirow{7}{*}{$\mathcal{T}_{\text{def}}$}} &
  Cars &
  \textit{NA} &
  \begin{tabular}[c]{@{}c@{}}73.34/92.97\\ 0.42/2.21\end{tabular} &
  \begin{tabular}[c]{@{}c@{}}73.36/99.28\\ 0.47/8.22\end{tabular} &
  \begin{tabular}[c]{@{}c@{}}71.81/95.33\\ 0.51/9.45\end{tabular} &
  \begin{tabular}[c]{@{}c@{}}73.62/96.90\\ 0.61/2.38\end{tabular} &
  \begin{tabular}[c]{@{}c@{}}72.89/99.43\\ 0.44/9.81\end{tabular} &
  \begin{tabular}[c]{@{}c@{}}73.78/71.81\\ 0.63/1.81\end{tabular} \\ \cline{2-9} 
\multicolumn{1}{c|}{} &
  RESISC45 &
  \begin{tabular}[c]{@{}c@{}}92.97/73.34\\ 2.10/0.56\end{tabular} &
  \textit{NA} &
  \begin{tabular}[c]{@{}c@{}}87.52/98.91\\ 2.14/4.56\end{tabular} &
  \begin{tabular}[c]{@{}c@{}}90.92/94.55\\ 2.37/7.58\end{tabular} &
  \begin{tabular}[c]{@{}c@{}}91.06/95.29\\ 1.94/1.96\end{tabular} &
  \begin{tabular}[c]{@{}c@{}}90.95/99.45\\ 2.51/9.41\end{tabular} &
  \begin{tabular}[c]{@{}c@{}}91.94/69.63\\ 2.27/2.82\end{tabular} \\ \cline{2-9} 
\multicolumn{1}{c|}{} &
  EuroSAT &
  \begin{tabular}[c]{@{}c@{}}99.28/73.36\\ 12.57/0.58\end{tabular} &
  \begin{tabular}[c]{@{}c@{}}98.91/87.52\\ 11.41/2.86\end{tabular} &
  \textit{NA} &
  \begin{tabular}[c]{@{}c@{}}97.65/92.96\\ 9.26/7.32\end{tabular} &
  \begin{tabular}[c]{@{}c@{}}97.19/92.14\\ 6.59/2.83\end{tabular} &
  \begin{tabular}[c]{@{}c@{}}97.93/99.20\\ 9.26/10.21\end{tabular} &
  \begin{tabular}[c]{@{}c@{}}99.07/68.99\\ 8.20/1.76\end{tabular} \\ \cline{2-9} 
\multicolumn{1}{c|}{} &
  SVHN &
  \begin{tabular}[c]{@{}c@{}}95.33/71.81\\ 6.69/0.47\end{tabular} &
  \begin{tabular}[c]{@{}c@{}}94.55/90.92\\ 7.54/2.52\end{tabular} &
  \begin{tabular}[c]{@{}c@{}}92.96/97.65\\ 9.56/10.76\end{tabular} &
  \textit{NA} &
  \begin{tabular}[c]{@{}c@{}}94.70/94.81\\ 14.11/3.10\end{tabular} &
  \begin{tabular}[c]{@{}c@{}}92.39/99.34\\ 8.75/9.50\end{tabular} &
  \begin{tabular}[c]{@{}c@{}}95.39/70.32\\ 7.76/2.02\end{tabular} \\ \cline{2-9} 
\multicolumn{1}{c|}{} &
  GTSRB &
  \begin{tabular}[c]{@{}c@{}}96.90/73.62\\ 2.71/0.56\end{tabular} &
  \begin{tabular}[c]{@{}c@{}}95.29/91.06\\ 2.13/2.10\end{tabular} &
  \begin{tabular}[c]{@{}c@{}}92.14/97.19\\ 2.97/7.94\end{tabular} &
  \begin{tabular}[c]{@{}c@{}}94.81/94.70\\ 2.12/7.58\end{tabular} &
  \textit{NA} &
  \begin{tabular}[c]{@{}c@{}}93.26/99.09\\ 3.44/11.31\end{tabular} &
  \begin{tabular}[c]{@{}c@{}}96.75/70.69\\ 2.15/1.60\end{tabular} \\ \cline{2-9} 
\multicolumn{1}{c|}{} &
  MNIST &
  \begin{tabular}[c]{@{}c@{}}99.43/72.89\\ 10.44/0.52\end{tabular} &
  \begin{tabular}[c]{@{}c@{}}99.45/90.95\\ 9.67/2.51\end{tabular} &
  \begin{tabular}[c]{@{}c@{}}99.20/97.93\\ 9.05/11.22\end{tabular} &
  \begin{tabular}[c]{@{}c@{}}99.34/92.39\\ 9.31/8.23\end{tabular} &
  \begin{tabular}[c]{@{}c@{}}99.09/93.26\\ 9.72/2.92\end{tabular} &
  \textit{NA} &
  \begin{tabular}[c]{@{}c@{}}99.45/68.83\\ 8.04/2.93\end{tabular} \\ \cline{2-9} 
\multicolumn{1}{c|}{} &
  DTD &
  \begin{tabular}[c]{@{}c@{}}71.81/73.78\\ 2.18/0.51\end{tabular} &
  \begin{tabular}[c]{@{}c@{}}69.63/91.94\\ 2.45/2.62\end{tabular} &
  \begin{tabular}[c]{@{}c@{}}68.99/99.07\\ 2.23/6.65\end{tabular} &
  \begin{tabular}[c]{@{}c@{}}70.32/95.39\\ 2.34/8.68\end{tabular} &
  \begin{tabular}[c]{@{}c@{}}70.69/96.75\\ 1.97/1.19\end{tabular} &
  \begin{tabular}[c]{@{}c@{}}68.83/99.45\\ 2.02/9.73\end{tabular} &
  \textit{NA} \\ \hline
\end{tabular}%
}
\end{table*}

\begin{table*}[t]
\centering
\caption{The classification accuracy of MMP-/MMP+ (merged by TIES) on $\mathcal{T}_{\text{def}}$ and $\mathcal{T}_{\text{fr}}$ on ViT-B-32.}
\label{tadare_vitb32_cpr}
\resizebox{\textwidth}{!}{%
\begin{tabular}{cc|ccccccc}
\hline
\multicolumn{2}{l|}{\multirow{2}{*}{\begin{tabular}[c]{@{}l@{}}MMP- Accuracy (\%) on $\mathcal{T}_{\text{def}}$/$\mathcal{T}_{\text{fr}}$\\ MMP+ Accuracy (\%) on $\mathcal{T}_{\text{def}}$/$\mathcal{T}_{\text{fr}}$\end{tabular}}} &
  \multicolumn{7}{c}{$\mathcal{T}_{\text{fr}}$} \\ \cline{3-9} 
\multicolumn{2}{l|}{} &
  Cars &
  RESISC45 &
  EuroSAT &
  SVHN &
  GTSRB &
  MNIST &
  DTD \\ \hline
\multicolumn{1}{c|}{\multirow{7}{*}{$\mathcal{T}_{\text{def}}$}} &
  Cars &
  \textit{NA} &
  \begin{tabular}[c]{@{}c@{}}75.96/93.35\\ 0.53/2.11\end{tabular} &
  \begin{tabular}[c]{@{}c@{}}76.11/99.11\\ 0.65/9.24\end{tabular} &
  \begin{tabular}[c]{@{}c@{}}75.28/96.27\\ 0.58/9.22\end{tabular} &
  \begin{tabular}[c]{@{}c@{}}75.95/96.60\\ 0.58/3.09\end{tabular} &
  \begin{tabular}[c]{@{}c@{}}76.43/99.57\\ 0.57/8.92\end{tabular} &
  \begin{tabular}[c]{@{}c@{}}75.46/74.57\\ 0.62/2.18\end{tabular} \\ \cline{2-9} 
\multicolumn{1}{c|}{} &
  RESISC45 &
  \begin{tabular}[c]{@{}c@{}}93.35/75.96\\ 3.22/0.52\end{tabular} &
  \textit{NA} &
  \begin{tabular}[c]{@{}c@{}}89.44/98.41\\ 2.75/19.61\end{tabular} &
  \begin{tabular}[c]{@{}c@{}}92.27/96.10\\ 2.86/7.85\end{tabular} &
  \begin{tabular}[c]{@{}c@{}}92.86/95.76\\ 3.05/3.08\end{tabular} &
  \begin{tabular}[c]{@{}c@{}}92.62/99.60\\ 3.02/9.85\end{tabular} &
  \begin{tabular}[c]{@{}c@{}}92.78/73.56\\ 3.16/2.29\end{tabular} \\ \cline{2-9} 
\multicolumn{1}{c|}{} &
  EuroSAT &
  \begin{tabular}[c]{@{}c@{}}99.11/76.11\\ 11.19/0.58\end{tabular} &
  \begin{tabular}[c]{@{}c@{}}98.41/89.44\\ 12.37/1.84\end{tabular} &
  \textit{NA} &
  \begin{tabular}[c]{@{}c@{}}96.96/95.39\\ 11.24/7.53\end{tabular} &
  \begin{tabular}[c]{@{}c@{}}96.87/92.61\\ 12.26/3.79\end{tabular} &
  \begin{tabular}[c]{@{}c@{}}98.15/99.51\\ 11.39/8.33\end{tabular} &
  \begin{tabular}[c]{@{}c@{}}98.63/72.93\\ 11.93/2.13\end{tabular} \\ \cline{2-9} 
\multicolumn{1}{c|}{} &
  SVHN &
  \begin{tabular}[c]{@{}c@{}}96.27/75.28\\ 9.18/0.47\end{tabular} &
  \begin{tabular}[c]{@{}c@{}}96.10/92.27\\ 9.17/2.11\end{tabular} &
  \begin{tabular}[c]{@{}c@{}}95.39/96.96\\ 9.16/10.93\end{tabular} &
  \textit{NA} &
  \begin{tabular}[c]{@{}c@{}}95.56/95.17\\ 9.18/4.81\end{tabular} &
  \begin{tabular}[c]{@{}c@{}}93.89/99.36\\ 9.16/8.92\end{tabular} &
  \begin{tabular}[c]{@{}c@{}}96.43/73.09\\ 9.16/2.45\end{tabular} \\ \cline{2-9} 
\multicolumn{1}{c|}{} &
  GTSRB &
  \begin{tabular}[c]{@{}c@{}}96.60/75.95\\ 2.95/0.53\end{tabular} &
  \begin{tabular}[c]{@{}c@{}}95.76/92.86\\ 3.00/3.41\end{tabular} &
  \begin{tabular}[c]{@{}c@{}}92.61/96.87\\ 2.96/9.19\end{tabular} &
  \begin{tabular}[c]{@{}c@{}}95.17/95.56\\ 2.90/9.56\end{tabular} &
  \textit{NA} &
  \begin{tabular}[c]{@{}c@{}}94.61/99.44\\ 3.01/10.02\end{tabular} &
  \begin{tabular}[c]{@{}c@{}}97.17/74.52\\ 2.89/1.86\end{tabular} \\ \cline{2-9} 
\multicolumn{1}{c|}{} &
  MNIST &
  \begin{tabular}[c]{@{}c@{}}99.57/76.43\\ 9.82/0.58\end{tabular} &
  \begin{tabular}[c]{@{}c@{}}99.60/92.62\\ 9.82/2.52\end{tabular} &
  \begin{tabular}[c]{@{}c@{}}99.51/98.15\\ 9.82/10.13\end{tabular} &
  \begin{tabular}[c]{@{}c@{}}99.36/93.89\\ 9.82/9.69\end{tabular} &
  \begin{tabular}[c]{@{}c@{}}99.44/94.61\\ 9.82/2.88\end{tabular} &
  \textit{NA} &
  \begin{tabular}[c]{@{}c@{}}99.59/72.82\\ 9.82/2.13\end{tabular} \\ \cline{2-9} 
\multicolumn{1}{c|}{} &
  DTD &
  \begin{tabular}[c]{@{}c@{}}74.57/75.46\\ 2.13/0.60\end{tabular} &
  \begin{tabular}[c]{@{}c@{}}73.56/92.78\\ 2.13/2.43\end{tabular} &
  \begin{tabular}[c]{@{}c@{}}72.93/98.63\\ 2.13/18.54\end{tabular} &
  \begin{tabular}[c]{@{}c@{}}73.09/96.43\\ 2.13/6.43\end{tabular} &
  \begin{tabular}[c]{@{}c@{}}74.52/97.17\\ 2.13/3.09\end{tabular} &
  \begin{tabular}[c]{@{}c@{}}72.82/99.59\\ 2.13/9.74\end{tabular} &
  \textit{NA} \\ \hline
\end{tabular}%
}
\end{table*}

\begin{table*}[t]
\centering
\caption{The classification accuracy of MMP-/MMP+ (merged by TIES with DARE) on $\mathcal{T}_{\text{def}}$ and $\mathcal{T}_{\text{fr}}$ on ViT-B-32.}
\label{tadare_vitb32_cpr}
\resizebox{\textwidth}{!}{%
\begin{tabular}{cc|ccccccc}
\hline
\multicolumn{2}{l|}{\multirow{2}{*}{\begin{tabular}[c]{@{}l@{}}MMP- Accuracy (\%) on $\mathcal{T}_{\text{def}}$/$\mathcal{T}_{\text{fr}}$\\ MMP+ Accuracy (\%) on $\mathcal{T}_{\text{def}}$/$\mathcal{T}_{\text{fr}}$\end{tabular}}} &
  \multicolumn{7}{c}{$\mathcal{T}_{\text{fr}}$} \\ \cline{3-9} 
\multicolumn{2}{l|}{} &
  Cars &
  RESISC45 &
  EuroSAT &
  SVHN &
  GTSRB &
  MNIST &
  DTD \\ \hline
\multicolumn{1}{c|}{\multirow{7}{*}{$\mathcal{T}_{\text{def}}$}} &
  Cars &
  \textit{NA} &
  \begin{tabular}[c]{@{}c@{}}75.76/90.84\\ 0.57/2.40\end{tabular} &
  \begin{tabular}[c]{@{}c@{}}76.16/98.78\\ 0.50/11.69\end{tabular} &
  \begin{tabular}[c]{@{}c@{}}75.67/95.44\\ 0.57/8.83\end{tabular} &
  \begin{tabular}[c]{@{}c@{}}76.12/95.55\\ 0.46/2.36\end{tabular} &
  \begin{tabular}[c]{@{}c@{}}76.32/99.53\\ 0.51/10.33\end{tabular} &
  \begin{tabular}[c]{@{}c@{}}75.03/71.33\\ 0.71/2.07\end{tabular} \\ \cline{2-9} 
\multicolumn{1}{c|}{} &
  RESISC45 &
  \begin{tabular}[c]{@{}c@{}}90.84/75.76\\ 2.13/0.57\end{tabular} &
  \textit{NA} &
  \begin{tabular}[c]{@{}c@{}}92.63/97.13\\ 2.70/10.81\end{tabular} &
  \begin{tabular}[c]{@{}c@{}}94.89/93.90\\ 2.11/7.55\end{tabular} &
  \begin{tabular}[c]{@{}c@{}}95.06/89.15\\ 2.10/2.08\end{tabular} &
  \begin{tabular}[c]{@{}c@{}}95.02/99.54\\ 2.21/9.73\end{tabular} &
  \begin{tabular}[c]{@{}c@{}}95.30/66.17\\ 2.05/2.18\end{tabular} \\ \cline{2-9} 
\multicolumn{1}{c|}{} &
  EuroSAT &
  \begin{tabular}[c]{@{}c@{}}98.78/76.16\\ 11.76/0.47\end{tabular} &
  \begin{tabular}[c]{@{}c@{}}97.13/92.63\\ 9.09/3.05\end{tabular} &
  \textit{NA} &
  \begin{tabular}[c]{@{}c@{}}99.50/89.06\\ 9.06/8.95\end{tabular} &
  \begin{tabular}[c]{@{}c@{}}99.56/82.42\\ 11.41/1.35\end{tabular} &
  \begin{tabular}[c]{@{}c@{}}99.65/99.12\\ 5.57/9.03\end{tabular} &
  \begin{tabular}[c]{@{}c@{}}99.81/64.47\\ 7.48/2.13\end{tabular} \\ \cline{2-9} 
\multicolumn{1}{c|}{} &
  SVHN &
  \begin{tabular}[c]{@{}c@{}}95.44/75.67\\ 12.92/0.53\end{tabular} &
  \begin{tabular}[c]{@{}c@{}}93.90/94.89\\ 9.12/2.51\end{tabular} &
  \begin{tabular}[c]{@{}c@{}}89.06/99.50\\ 8.53/9.39\end{tabular} &
  \textit{NA} &
  \begin{tabular}[c]{@{}c@{}}96.81/87.63\\ 8.92/2.08\end{tabular} &
  \begin{tabular}[c]{@{}c@{}}96.25/99.05\\ 10.11/10.10\end{tabular} &
  \begin{tabular}[c]{@{}c@{}}97.26/62.45\\ 8.77/1.86\end{tabular} \\ \cline{2-9} 
\multicolumn{1}{c|}{} &
  GTSRB &
  \begin{tabular}[c]{@{}c@{}}95.55/76.12\\ 2.58/0.51\end{tabular} &
  \begin{tabular}[c]{@{}c@{}}89.15/95.06\\ 1.80/1.97\end{tabular} &
  \begin{tabular}[c]{@{}c@{}}82.42/99.56\\ 0.80/7.83\end{tabular} &
  \begin{tabular}[c]{@{}c@{}}87.63/96.81\\ 1.94/8.84\end{tabular} &
  \textit{NA} &
  \begin{tabular}[c]{@{}c@{}}98.27/98.63\\ 2.53/10.73\end{tabular} &
  \begin{tabular}[c]{@{}c@{}}98.73/61.70\\ 1.38/2.77\end{tabular} \\ \cline{2-9} 
\multicolumn{1}{c|}{} &
  MNIST &
  \begin{tabular}[c]{@{}c@{}}99.53/76.32\\ 9.62/0.49\end{tabular} &
  \begin{tabular}[c]{@{}c@{}}99.54/95.02\\ 9.16/1.98\end{tabular} &
  \begin{tabular}[c]{@{}c@{}}99.12/99.65\\ 11.07/5.93\end{tabular} &
  \begin{tabular}[c]{@{}c@{}}99.05/96.25\\ 9.48/8.03\end{tabular} &
  \begin{tabular}[c]{@{}c@{}}98.63/98.27\\ 10.52/2.06\end{tabular} &
  \textit{NA} &
  \begin{tabular}[c]{@{}c@{}}99.67/60.53\\ 9.33/2.29\end{tabular} \\ \cline{2-9} 
\multicolumn{1}{c|}{} &
  DTD &
  \begin{tabular}[c]{@{}c@{}}71.33/75.03\\ 2.02/0.47\end{tabular} &
  \begin{tabular}[c]{@{}c@{}}66.17/95.30\\ 2.45/2.56\end{tabular} &
  \begin{tabular}[c]{@{}c@{}}64.47/99.81\\ 2.82/11.98\end{tabular} &
  \begin{tabular}[c]{@{}c@{}}62.45/97.26\\ 2.18/8.31\end{tabular} &
  \begin{tabular}[c]{@{}c@{}}61.70/98.73\\ 1.70/2.15\end{tabular} &
  \begin{tabular}[c]{@{}c@{}}60.53/99.67\\ 2.02/8.79\end{tabular} &
  \textit{NA} \\ \hline
\end{tabular}%
}
\end{table*}

\begin{table*}[t]
\centering
\caption{The classification accuracy of MMP-/MMP+ (merged by AdaMerging) on $\mathcal{T}_{\text{def}}$ and $\mathcal{T}_{\text{fr}}$ on ViT-B-32.}
\label{adamerging_vitb32_cpr}
\resizebox{\textwidth}{!}{%
\begin{tabular}{cc|ccccccc}
\hline
\multicolumn{2}{l|}{\multirow{2}{*}{\begin{tabular}[c]{@{}l@{}}MMP- Accuracy (\%) on $\mathcal{T}_{\text{def}}$/$\mathcal{T}_{\text{fr}}$\\ MMP+ Accuracy (\%) on $\mathcal{T}_{\text{def}}$/$\mathcal{T}_{\text{fr}}$\end{tabular}}} &
  \multicolumn{7}{c}{$\mathcal{T}_{\text{fr}}$} \\ \cline{3-9} 
\multicolumn{2}{l|}{} &
  Cars &
  RESISC45 &
  EuroSAT &
  SVHN &
  GTSRB &
  MNIST &
  DTD \\ \hline
\multicolumn{1}{c|}{\multirow{7}{*}{$\mathcal{T}_{\text{def}}$}} &
  Cars &
  \textit{NA} &
  \begin{tabular}[c]{@{}c@{}}74.01/90.67\\ 0.61/2.52\end{tabular} &
  \begin{tabular}[c]{@{}c@{}}73.72/96.81\\ 0.57/21.74\end{tabular} &
  \begin{tabular}[c]{@{}c@{}}73.51/83.82\\ 0.56/11.25\end{tabular} &
  \begin{tabular}[c]{@{}c@{}}73.71/95.69\\ 0.55/3.04\end{tabular} &
  \begin{tabular}[c]{@{}c@{}}74.56/96.51\\ 0.57/9.68\end{tabular} &
  \begin{tabular}[c]{@{}c@{}}73.80/66.12\\ 0.67/2.23\end{tabular} \\ \cline{2-9} 
\multicolumn{1}{c|}{} &
  RESISC45 &
  \begin{tabular}[c]{@{}c@{}}90.67/74.01\\ 2.81/0.55\end{tabular} &
  \textit{NA} &
  \begin{tabular}[c]{@{}c@{}}87.29/97.09\\ 2.60/13.06\end{tabular} &
  \begin{tabular}[c]{@{}c@{}}90.29/83.99\\ 2.57/7.64\end{tabular} &
  \begin{tabular}[c]{@{}c@{}}89.10/95.46\\ 2.90/2.14\end{tabular} &
  \begin{tabular}[c]{@{}c@{}}90.51/96.77\\ 2.68/9.51\end{tabular} &
  \begin{tabular}[c]{@{}c@{}}90.59/66.17\\ 2.84/2.93\end{tabular} \\ \cline{2-9} 
\multicolumn{1}{c|}{} &
  EuroSAT &
  \begin{tabular}[c]{@{}c@{}}96.81/73.72\\ 9.26/0.51\end{tabular} &
  \begin{tabular}[c]{@{}c@{}}97.09/87.29\\ 9.26/2.52\end{tabular} &
  \textit{NA} &
  \begin{tabular}[c]{@{}c@{}}95.91/82.77\\ 9.20/11.74\end{tabular} &
  \begin{tabular}[c]{@{}c@{}}95.00/95.15\\ 9.24/3.80\end{tabular} &
  \begin{tabular}[c]{@{}c@{}}96.50/96.32\\ 9.26/8.35\end{tabular} &
  \begin{tabular}[c]{@{}c@{}}96.76/66.33\\ 9.26/1.12\end{tabular} \\ \cline{2-9} 
\multicolumn{1}{c|}{} &
  SVHN &
  \begin{tabular}[c]{@{}c@{}}83.82/73.51\\ 7.82/0.55\end{tabular} &
  \begin{tabular}[c]{@{}c@{}}83.99/90.29\\ 7.70/2.43\end{tabular} &
  \begin{tabular}[c]{@{}c@{}}82.77/95.91\\ 7.68/13.17\end{tabular} &
  \textit{NA} &
  \begin{tabular}[c]{@{}c@{}}87.63/95.91\\ 7.94/3.09\end{tabular} &
  \begin{tabular}[c]{@{}c@{}}87.77/98.22\\ 7.79/9.58\end{tabular} &
  \begin{tabular}[c]{@{}c@{}}85.15/66.81\\ 7.64/2.29\end{tabular} \\ \cline{2-9} 
\multicolumn{1}{c|}{} &
  GTSRB &
  \begin{tabular}[c]{@{}c@{}}95.69/73.71\\ 1.46/0.55\end{tabular} &
  \begin{tabular}[c]{@{}c@{}}95.46/89.10\\ 1.70/3.16\end{tabular} &
  \begin{tabular}[c]{@{}c@{}}95.15/95.00\\ 1.49/11.41\end{tabular} &
  \begin{tabular}[c]{@{}c@{}}95.91/87.63\\ 1.54/8.10\end{tabular} &
  \textit{NA} &
  \begin{tabular}[c]{@{}c@{}}95.72/96.58\\ 1.47/9.97\end{tabular} &
  \begin{tabular}[c]{@{}c@{}}95.66/66.54\\ 1.74/2.13\end{tabular} \\ \cline{2-9} 
\multicolumn{1}{c|}{} &
  MNIST &
  \begin{tabular}[c]{@{}c@{}}96.51/74.56\\ 6.68/0.44\end{tabular} &
  \begin{tabular}[c]{@{}c@{}}96.77/90.51\\ 5.32/2.78\end{tabular} &
  \begin{tabular}[c]{@{}c@{}}96.32/96.50\\ 6.84/11.81\end{tabular} &
  \begin{tabular}[c]{@{}c@{}}98.22/87.77\\ 7.53/19.11\end{tabular} &
  \begin{tabular}[c]{@{}c@{}}96.58/95.72\\ 8.28/6.03\end{tabular} &
  \textit{NA} &
  \begin{tabular}[c]{@{}c@{}}96.73/66.38\\ 8.37/2.18\end{tabular} \\ \cline{2-9} 
\multicolumn{1}{c|}{} &
  DTD &
  \begin{tabular}[c]{@{}c@{}}66.12/73.80\\ 1.91/0.36\end{tabular} &
  \begin{tabular}[c]{@{}c@{}}66.17/90.59\\ 2.07/4.00\end{tabular} &
  \begin{tabular}[c]{@{}c@{}}66.33/96.76\\ 1.97/8.41\end{tabular} &
  \begin{tabular}[c]{@{}c@{}}66.81/85.15\\ 2.13/12.69\end{tabular} &
  \begin{tabular}[c]{@{}c@{}}66.54/95.66\\ 2.34/3.09\end{tabular} &
  \begin{tabular}[c]{@{}c@{}}66.38/96.73\\ 1.86/8.92\end{tabular} &
  \textit{NA} \\ \hline
\end{tabular}%
}
\end{table*}

\begin{table*}[t]
\centering
\caption{The classification accuracy of MMP-/MMP+ (merged by AdaMerging with DARE) on $\mathcal{T}_{\text{def}}$ and $\mathcal{T}_{\text{fr}}$ on ViT-B-32.}
\label{adamerging_dare_vitb32_cpr}
\resizebox{\textwidth}{!}{%
\begin{tabular}{cc|ccccccc}
\hline
\multicolumn{2}{l|}{\multirow{2}{*}{\begin{tabular}[c]{@{}l@{}}MMP- Accuracy (\%) on $\mathcal{T}_{\text{def}}$/$\mathcal{T}_{\text{fr}}$\\ MMP+ Accuracy (\%) on $\mathcal{T}_{\text{def}}$/$\mathcal{T}_{\text{fr}}$\end{tabular}}} &
  \multicolumn{7}{c}{$\mathcal{T}_{\text{fr}}$} \\ \cline{3-9} 
\multicolumn{2}{l|}{} &
  Cars &
  RESISC45 &
  EuroSAT &
  SVHN &
  GTSRB &
  MNIST &
  DTD \\ \hline
\multicolumn{1}{c|}{\multirow{7}{*}{$\mathcal{T}_{\text{def}}$}} &
  Cars &
  \textit{NA} &
  \begin{tabular}[c]{@{}c@{}}73.85/90.41\\ 0.51/1.87\end{tabular} &
  \begin{tabular}[c]{@{}c@{}}73.90/96.83\\ 0.56/9.28\end{tabular} &
  \begin{tabular}[c]{@{}c@{}}73.27/83.97\\ 0.50/8.70\end{tabular} &
  \begin{tabular}[c]{@{}c@{}}73.68/95.63\\ 0.34/3.06\end{tabular} &
  \begin{tabular}[c]{@{}c@{}}74.37/96.55\\ 0.47/11.63\end{tabular} &
  \begin{tabular}[c]{@{}c@{}}73.97/66.33\\ 0.49/2.34\end{tabular} \\ \cline{2-9} 
\multicolumn{1}{c|}{} &
  RESISC45 &
  \begin{tabular}[c]{@{}c@{}}90.41/73.85\\ 1.94/0.50\end{tabular} &
  \textit{NA} &
  \begin{tabular}[c]{@{}c@{}}87.30/96.91\\ 2.33/9.15\end{tabular} &
  \begin{tabular}[c]{@{}c@{}}90.30/83.87\\ 2.60/9.69\end{tabular} &
  \begin{tabular}[c]{@{}c@{}}89.10/95.49\\ 2.40/3.26\end{tabular} &
  \begin{tabular}[c]{@{}c@{}}90.32/96.72\\ 2.02/9.63\end{tabular} &
  \begin{tabular}[c]{@{}c@{}}90.70/66.38\\ 3.17/2.61\end{tabular} \\ \cline{2-9} 
\multicolumn{1}{c|}{} &
  EuroSAT &
  \begin{tabular}[c]{@{}c@{}}96.83/73.90\\ 13.59/0.51\end{tabular} &
  \begin{tabular}[c]{@{}c@{}}96.91/87.30\\ 9.48/1.97\end{tabular} &
  \textit{NA} &
  \begin{tabular}[c]{@{}c@{}}95.89/83.01\\ 13.43/13.14\end{tabular} &
  \begin{tabular}[c]{@{}c@{}}94.94/94.96\\ 11.30/1.84\end{tabular} &
  \begin{tabular}[c]{@{}c@{}}96.56/96.40\\ 7.91/10.28\end{tabular} &
  \begin{tabular}[c]{@{}c@{}}96.70/66.12\\ 15.44/1.97\end{tabular} \\ \cline{2-9} 
\multicolumn{1}{c|}{} &
  SVHN &
  \begin{tabular}[c]{@{}c@{}}83.97/73.27\\ 9.70/0.50\end{tabular} &
  \begin{tabular}[c]{@{}c@{}}83.87/90.30\\ 8.09/2.19\end{tabular} &
  \begin{tabular}[c]{@{}c@{}}83.01/95.89\\ 9.75/11.30\end{tabular} &
  \textit{NA} &
  \begin{tabular}[c]{@{}c@{}}87.80/95.95\\ 9.15/1.36\end{tabular} &
  \begin{tabular}[c]{@{}c@{}}87.83/98.21\\ 7.24/11.10\end{tabular} &
  \begin{tabular}[c]{@{}c@{}}84.95/66.97\\ 8.55/1.86\end{tabular} \\ \cline{2-9} 
\multicolumn{1}{c|}{} &
  GTSRB &
  \begin{tabular}[c]{@{}c@{}}95.63/73.68\\ 1.59/0.56\end{tabular} &
  \begin{tabular}[c]{@{}c@{}}95.49/89.10\\ 1.36/1.81\end{tabular} &
  \begin{tabular}[c]{@{}c@{}}94.96/94.94\\ 3.09/9.33\end{tabular} &
  \begin{tabular}[c]{@{}c@{}}95.95/87.80\\ 2.12/7.62\end{tabular} &
  \textit{NA} &
  \begin{tabular}[c]{@{}c@{}}95.62/96.62\\ 1.76/9.40\end{tabular} &
  \begin{tabular}[c]{@{}c@{}}95.61/66.44\\ 2.89/3.09\end{tabular} \\ \cline{2-9} 
\multicolumn{1}{c|}{} &
  MNIST &
  \begin{tabular}[c]{@{}c@{}}96.55/74.37\\ 10.28/0.46\end{tabular} &
  \begin{tabular}[c]{@{}c@{}}96.72/90.32\\ 9.82/1.17\end{tabular} &
  \begin{tabular}[c]{@{}c@{}}96.40/96.56\\ 10.10/11.11\end{tabular} &
  \begin{tabular}[c]{@{}c@{}}98.21/87.83\\ 8.51/7.99\end{tabular} &
  \begin{tabular}[c]{@{}c@{}}96.62/95.62\\ 10.12/2.16\end{tabular} &
  \textit{NA} &
  \begin{tabular}[c]{@{}c@{}}96.77/66.38\\ 9.52/1.44\end{tabular} \\ \cline{2-9} 
\multicolumn{1}{c|}{} &
  DTD &
  \begin{tabular}[c]{@{}c@{}}66.33/73.97\\ 2.23/0.44\end{tabular} &
  \begin{tabular}[c]{@{}c@{}}66.38/90.70\\ 2.29/2.11\end{tabular} &
  \begin{tabular}[c]{@{}c@{}}66.12/96.70\\ 2.13/9.04\end{tabular} &
  \begin{tabular}[c]{@{}c@{}}66.97/84.95\\ 2.13/9.04\end{tabular} &
  \begin{tabular}[c]{@{}c@{}}66.44/95.61\\ 2.29/2.57\end{tabular} &
  \begin{tabular}[c]{@{}c@{}}66.38/96.77\\ 2.39/9.10\end{tabular} &
  \textit{NA} \\ \hline
\end{tabular}%
}
\end{table*}

\end{document}